\documentclass[10pt,journal,compsoc,twoside]{IEEEtran}
\ifCLASSOPTIONcompsoc
  \usepackage[nocompress]{cite}
\else
  \usepackage{cite}
\fi
\ifCLASSINFOpdf
\else
\fi
\usepackage{times}
\usepackage{epsfig}
\usepackage{graphicx}
\usepackage{amsmath}
\usepackage{amssymb}
\usepackage{algorithm,algpseudocode}

\graphicspath{{Figures/}}

\usepackage{makecell}

\usepackage{amsmath,amssymb,amsfonts} 
\usepackage{subcaption}
\usepackage{fixltx2e}
\usepackage{multirow}
\usepackage{booktabs}
\usepackage{amsmath}

\usepackage{algorithm,algpseudocode}

\algrenewcommand\algorithmicforall{\textbf{foreach}}

\algdef{SE}[SUBALG]{SubAlgorithm}{EndSubAlgorithm}{\algorithmicsubalgorithm}{\algorithmicend\ \algorithmicsubalgorithm}%

\algtext*{SubAlgorithm}
\algtext*{EndSubAlgorithm}% If you want to avoid seeing "end sub-algorithm"

\algdef{SE}[SUBALG]{Indent}{EndIndent}{}{\algorithmicend\ }%
\algtext*{Indent}
\algtext*{EndIndent}

\usepackage{threeparttable}

\usepackage{arydshln}
\usepackage{scalerel}
\usepackage{tikz}
\usetikzlibrary{svg.path}
\definecolor{orcidlogocol}{HTML}{A6CE39}
\tikzset{
	orcidlogo/.pic={
		\fill[orcidlogocol] svg{M256,128c0,70.7-57.3,128-128,128C57.3,256,0,198.7,0,128C0,57.3,57.3,0,128,0C198.7,0,256,57.3,256,128z};
		\fill[white] svg{M86.3,186.2H70.9V79.1h15.4v48.4V186.2z}
		svg{M108.9,79.1h41.6c39.6,0,57,28.3,57,53.6c0,27.5-21.5,53.6-56.8,53.6h-41.8V79.1z M124.3,172.4h24.5c34.9,0,42.9-26.5,42.9-39.7c0-21.5-13.7-39.7-43.7-39.7h-23.7V172.4z}
		svg{M88.7,56.8c0,5.5-4.5,10.1-10.1,10.1c-5.6,0-10.1-4.6-10.1-10.1c0-5.6,4.5-10.1,10.1-10.1C84.2,46.7,88.7,51.3,88.7,56.8z};
	}
}

\newcommand\orcidicon[1]{\href{https://orcid.org/#1}{\mbox{\scalerel*{
				\begin{tikzpicture}[yscale=-1,transform shape]
					\pic{orcidlogo};
				\end{tikzpicture}
			}{|}}}}
		
\usepackage[hidelinks]{hyperref}
\usepackage{url}

\begin{document}
%
% paper title
% Titles are generally capitalized except for words such as a, an, and, as,
% at, but, by, for, in, nor, of, on, or, the, to and up, which are usually
% not capitalized unless they are the first or last word of the title.
% Linebreaks \\ can be used within to get better formatting as desired.
% Do not put math or special symbols in the title.
\title{Ranked List Loss for Deep Metric Learning}
%
%
% author names and IEEE memberships
% note positions of commas and nonbreaking spaces ( ~ ) LaTeX will not break
% a structure at a ~ so this keeps an author's name from being broken across
% two lines.
% use \thanks{} to gain access to the first footnote area
% a separate \thanks must be used for each paragraph as LaTeX2e's \thanks
% was not built to handle multiple paragraphs
%
%
%\IEEEcompsocitemizethanks is a special \thanks that produces the bulleted
% lists the Computer Society journals use for "first footnote" author
% affiliations. Use \IEEEcompsocthanksitem which works much like \item
% for each affiliation group. When not in compsoc mode,
% \IEEEcompsocitemizethanks becomes like \thanks and
% \IEEEcompsocthanksitem becomes a line break with idention. This
% facilitates dual compilation, although admittedly the differences in the
% desired content of \author between the different types of papers makes a
% one-size-fits-all approach a daunting prospect. For instance, compsoc 
% journal papers have the author affiliations above the "Manuscript
% received ..."  text while in non-compsoc journals this is reversed. Sigh.

\author{Xinshao~Wang$^{1,2,\textsuperscript{\orcidicon{0000-0001-8907-8258}}}$,%~\IEEEmembership{Member,~IEEE,}
        ~Yang~Hua$^{1, \textsuperscript{\orcidicon{0000-0001-5536-503X}}}$, %\IEEEmembership{Member, IEEE}
        ~Elyor~Kodirov,%~\IEEEmembership{Fellow,~OSA,}
        ~and~Neil~M.~Robertson$^{1}$, \IEEEmembership{Senior Member, IEEE}%~\IEEEmembership{Fellow,~OSA,}
        %and~Jane~Doe,~\IEEEmembership{Life~Fellow,~IEEE}% <-this % stops a space
\IEEEcompsocitemizethanks{
\IEEEcompsocthanksitem 
%	$^1$School of Electronics, Electrical Engineering and Computer Science, Queen's University Belfast, UK.
    $^1$EEECS/ECIT, Queen's University Belfast, UK.
	%\IEEEcompsocthanksitem 
	%$^2$Anyvision Research Team, UK.
	$^2$Department of Engineering Science, University of Oxford, UK.
%\IEEEcompsocthanksitem Corresponding author: \{y.hua\}@qub.ac.uk.
\IEEEcompsocthanksitem 
\{xinshao.wang\}@eng.ox.ac.uk,
\{y.hua,~n.robertson\}@qub.ac.uk. 
\IEEEcompsocthanksitem 
Corresponding author: Yang Hua. 
}% <-this % stops a space
%\thanks{Manuscript received August 19, 2019; revised August 19, 2019.}
}

% note the % following the last \IEEEmembership and also \thanks - 
% these prevent an unwanted space from occurring between the last author name
% and the end of the author line. i.e., if you had this:
% 
% \author{....lastname \thanks{...} \thanks{...} }
%                     ^------------^------------^----Do not want these spaces!
%
% a space would be appended to the last name and could cause every name on that
% line to be shifted left slightly. This is one of those "LaTeX things". For
% instance, "\textbf{A} \textbf{B}" will typeset as "A B" not "AB". To get
% "AB" then you have to do: "\textbf{A}\textbf{B}"
% \thanks is no different in this regard, so shield the last } of each \thanks
% that ends a line with a % and do not let a space in before the next \thanks.
% Spaces after \IEEEmembership other than the last one are OK (and needed) as
% you are supposed to have spaces between the names. For what it is worth,
% this is a minor point as most people would not even notice if the said evil
% space somehow managed to creep in.

% The paper headers
\markboth{
	IEEE Transactions on Pattern Analysis \& Machine Intelligence,~Vol.~xx, No.~xx, month~20xx
}{Wang {{et al.}}: Ranked List Loss for Deep Metric Learning}
% The only time the second header will appear is for the odd numbered pages
% after the title page when using the twoside option.
% 
% *** Note that you probably will NOT want to include the author's ***
% *** name in the headers of peer review papers.                   ***
% You can use \ifCLASSOPTIONpeerreview for conditional compilation here if
% you desire.

% The publisher's ID mark at the bottom of the page is less important with
% Computer Society journal papers as those publications place the marks
% outside of the main text columns and, therefore, unlike regular IEEE
% journals, the available text space is not reduced by their presence.
% If you want to put a publisher's ID mark on the page you can do it like
% this:
%\IEEEpubid{0000--0000/00\$00.00~\copyright~2015 IEEE}
% or like this to get the Computer Society new two part style.
%\IEEEpubid{\makebox[\columnwidth]{\hfill 0000--0000/00/\$00.00~\copyright~2015 IEEE}%
%\hspace{\columnsep}\makebox[\columnwidth]{Published by the IEEE Computer Society\hfill}}
% Remember, if you use this you must call \IEEEpubidadjcol in the second
% column for its text to clear the IEEEpubid mark (Computer Society journal
% papers don't need this extra clearance.)

% use for special paper notices
%\IEEEspecialpapernotice{(Invited Paper)}

% for Computer Society papers, we must declare the abstract and index terms
% PRIOR to the title within the \IEEEtitleabstractindextext IEEEtran
% command as these need to go into the title area created by \maketitle.
% As a general rule, do not put math, special symbols or citations
% in the abstract or keywords.
\IEEEtitleabstractindextext{%
\begin{abstract}
The objective of deep metric learning (DML) is to learn embeddings that can capture semantic similarity and dissimilarity information among data points. 
%DML has wide applications, such as face recognition and verification, image retrieval and clustering. 
%\\
%e
Existing pairwise or tripletwise loss functions used in DML are known to suffer from slow convergence due to a large proportion of trivial pairs or triplets as the model improves. 
To improve this, ranking-motivated structured losses are proposed recently to incorporate multiple examples and exploit the structured information among them. 
They converge faster and achieve state-of-the-art performance.
%Therefore, the structured losses which capture the similarity structure among multiple examples are widely studied recently. As more samples are incorporated and the structured information is exploited, they converges faster and achieves state-of-the-art performance.  
%
In this work, we unveil two limitations of existing ranking-motivated structured losses and propose a novel ranked list loss to solve both of them. 
First, given a query, only a fraction of data points is incorporated to build the similarity structure. Consequently, some useful examples are ignored and the structure is less informative. To address this, we propose to build a set-based similarity structure by exploiting all instances in the gallery. 
The learning setting can be interpreted as few-shot retrieval: given a mini-batch, every example is iteratively used as a query, and the rest ones compose the gallery to search, i.e., the support set in few-shot setting.  
The rest examples are split into a positive set and a negative set. For every mini-batch, the learning objective of ranked list loss is to make the query closer to the positive set than to the negative set by a margin.
Second, previous methods aim to pull positive pairs as close as possible in the embedding space.
% and thus compress each class into one point 
As a result, the intraclass data distribution tends to be extremely compressed. In contrast, we propose to learn a hypersphere for each class in order to preserve useful similarity structure inside it, which functions as regularisation.   
%
%
%% Result, performs on par with
Extensive experiments demonstrate the superiority of our proposal  by comparing with the state-of-the-art methods on the fine-grained image retrieval task.  
Our source code is available online: https://github.com/XinshaoAmos\\Wang/Ranked-List-Loss-for-DML. %\href{https://github.com/XinshaoAmosWang/Ranked-List-Loss-for-DML}{https://github.com/XinshaoAmos\\Wang/Ranked-List-Loss-for-DML}.
%in a variety of contexts including online shopping products, cars and birds.  
%i.e., CARS196 \cite{krause20133d}, CUB-200-2011 \cite{wah2011caltech}, and Stanford Online Products (SOP) \cite{song2016deep}. 
\end{abstract}

% Note that keywords are not normally used for peerreview papers.
}

% make the title area
\maketitle

% To allow for easy dual compilation without having to reenter the
% abstract/keywords data, the \IEEEtitleabstractindextext text will
% not be used in maketitle, but will appear (i.e., to be "transported")
% here as \IEEEdisplaynontitleabstractindextext when compsoc mode
% is not selected <OR> if conference mode is selected - because compsoc
% conference papers position the abstract like regular (non-compsoc)
% papers do!
\IEEEdisplaynontitleabstractindextext
% \IEEEdisplaynontitleabstractindextext has no effect when using
% compsoc under a non-conference mode.

% For peer review papers, you can put extra information on the cover
% page as needed:
% \ifCLASSOPTIONpeerreview
% \begin{center} \bfseries EDICS Category: 3-BBND \end{center}
% \fi
%
% For peerreview papers, this IEEEtran command inserts a page break and
% creates the second title. It will be ignored for other modes.
\IEEEpeerreviewmaketitle

\ifCLASSOPTIONcompsoc
\IEEEraisesectionheading{\section{Introduction}\label{sec:introduction}}
\else
\section{Introduction}
\label{sec:introduction}
\fi

\IEEEPARstart{D}{eep} metric learning (DML) plays a crucial role in a variety of applications in computer vision, such as image retrieval \cite{sohn2016improved,movshovitz2017no}, clustering \cite{hershey2016deep}, and transfer learning \cite{song2016deep}.
For example, by using DML, FaceNet \cite{schroff2015facenet} achieves superhuman performance on face verification with 260M face images of 8M identities. 
%Generally, the classification approaches become impractical in the cases with a prohibitively large label set.

%\emph{Loss Functions}\\
%\emph{Negative Pairs Sampling}\\
%\emph{Motivation} \\

Loss function 
%plays a key role
is one of essential components 
in successful DML frameworks and a large variety of loss functions have been proposed in the literature. 
%For example, 
Contrastive loss \cite{chopra2005learning,hadsell2006dimensionality} captures the relationship between pairwise data points, i.e., similarity or dissimilarity. Triplet-based losses are also widely studied \cite{schroff2015facenet,wang2014learning,cui2016fine}. A triplet is composed of an anchor point, a similar (positive) data point and dissimilar (negative) data point. The purpose of triplet loss is to learn a distance metric by which the anchor point is closer to the similar point than the dissimilar one by a margin.
In general, the triplet loss outperforms the contrastive loss \cite{song2016deep,schroff2015facenet} because the relationship between positive and negative pairs is considered. 
Inspired by this, recent ranking-motivated methods\footnote{We term them ranking-motivated methods, whose target is to make the largest distance of similar pairs smaller than the smallest distance of dissimilar pairs. In this context, we do not care the distance order inside positive and negative sets.
Namely, rank and retrieve can be used interchangeably here.} \cite{schroff2015facenet,sohn2016improved,song2016deep,song2017deep,law2017deep,movshovitz2017no} propose to take into consideration the richer structured information among multiple data points and achieve impressive performance on many applications, e.g., fine-grained image retrieval and clustering.
%Existing structured loss functions are presented in detail in the section~\ref{section:preliminary}. 

%(1) a separate data preparation stage to construct rigid pair format. \\
%(2) not making use of all pairs/information.
%
%One of the most widely used evaluation metric of DML is image retrieval performance (Recall@$K$). Given a query, the objective of image retrieval is to produce a list ranked by the similarity scores of the query and instances in the gallery. If images' embeddings are learned well, the matching instances should be ranked in the first $K$ positions. 
%
%

%%%%%%%%%%%%%%%%%%%%%%%%%%%%%%%%%%%%%%%%%%%%%%%%%%%%%
\begin{figure}[!t]
	\centering
	\includegraphics[width=\linewidth]{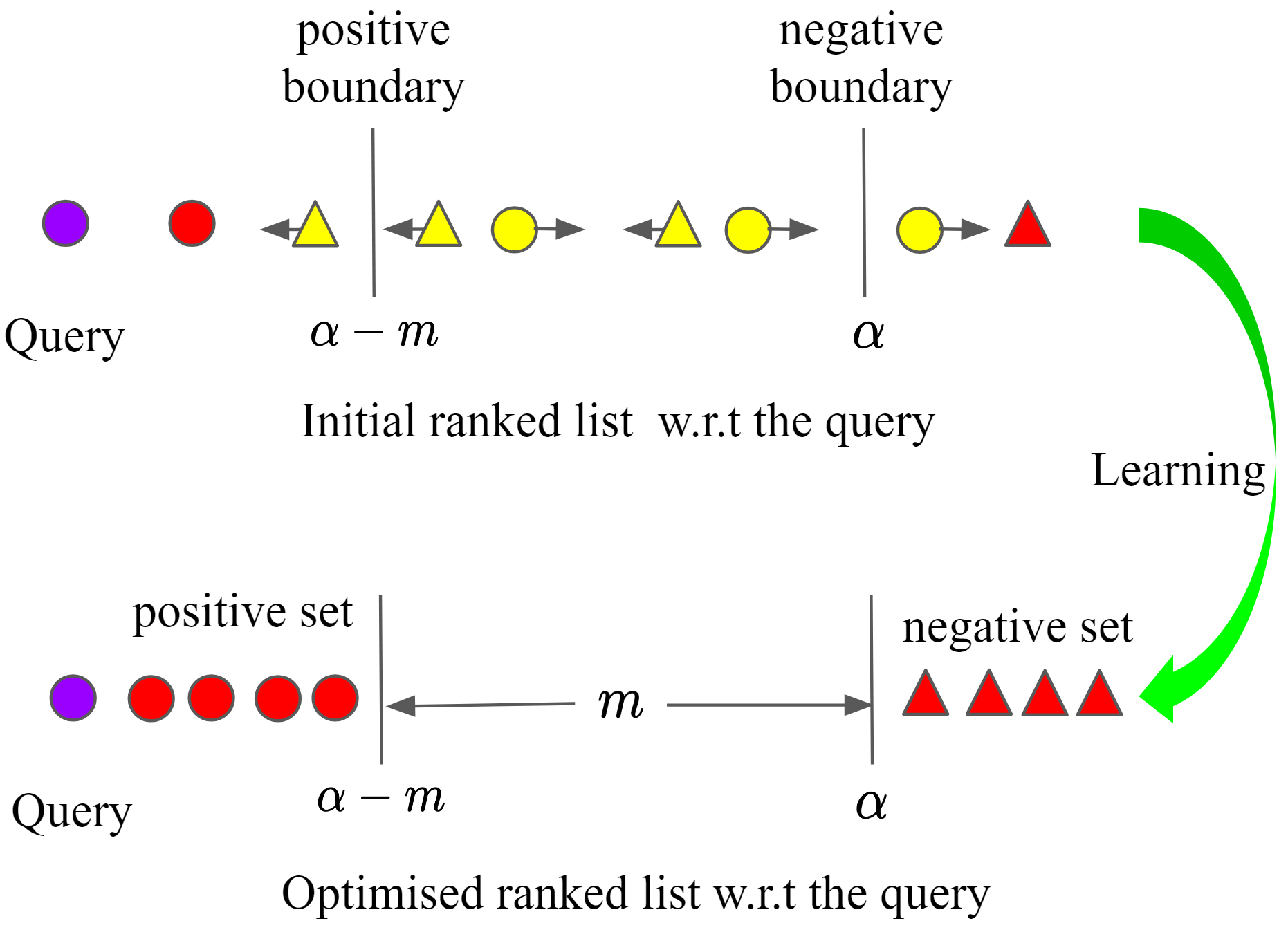}
	\caption{
		Illustration of our proposed Ranked List Loss (RLL). 
		Given a query and its ranked list, 
		RLL aims to make the query closer to the positive set than to the negative set by a margin $m$.   
		%optimises one query's embedding by considering all non-trivial positive and negative examples in the ranked list.
		%We compute the dual truncated contrastive loss of each query with all other images.
		Circle and triangle  represent two different classes. 
		The blue circle is a query. 
		The yellow shapes represent nontrivial examples while the red shapes represent trivial examples.
		%The red and yellow circles represents trivial and non-trivial positive examples respectively. The red and yellow triangles represents trivial and non-trivial negative instances respectively.   
		%
		The arrow indicates the query's gradient direction determined by the corresponding non-trivial examples. The final gradient direction of the query is {a weighted combination of them}.
		%\textit{Best viewed in colour.}
		The optimisation target for every list is shown in the bottom. \textit{Best viewed in colour.}  
	}
	\label{fig:ranked_list_loss}
	\vspace{-0.25cm}
\end{figure}
%%%%%%%%%%%%%%%%%%%%%%%%%%%%%%%%%%%%%%%%%%%%%%%%%%%%

However, there are still certain limitations in current state-of-the-art DML approaches.  
Firstly, we notice that \textit{only a proportion of informative examples} is incorporated to capture the structure in previous ranking-motivated 
%\note{need to explain it before use it} 
loss functions. In this case, some non-trivial examples are wasted and the structured information is extracted from fewer data points.
To address it, we propose to \textit{utilise all non-trivial data points} to build a more informative structure and exploit it to learn more discriminative embeddings. Specifically, given a query, we obtain a ranked list by sorting all other data points (gallery) according to the similarities. 
%\note{2 repetitions : query, ranking. Instead I propose here to say "Given a query, we obtain a ranked list by sorting all other data points (gallery) according to their similarities"} 
Ideally, all the positive examples are supposed to be ranked before the negative samples in the feature space.
%\note{feature space instead? To diversify vocabulary}. 
To achieve this, we introduce ranked list loss (RLL) to organise the samples of each query. Given a query, the optimisation of RLL is to rank all positive points before the negative points and forcing a margin between them. 
%\note{"the optimisation objective" not necessary instead "optimisation" is enough }.    
%As a result, the anchor is closer to the positive set than the negative set by a margin. 
In other words, RLL aims to explore the set-based similarity structure, which contains richer information than the point-based approach, e.g., triplet loss. 

%The objective of triplet loss is to pull the anchor closer to the positive point than to the negative point by a margin. 
%
%The proposed RLL can be also viewed as a \textit{set-based triplet loss}. In the ranked list of each query (anchor),  we pull the anchor closer to the positive set than to the negative set by a fixed margin. 
%we are \textit{the first to incorporate all non-trivial data points} to capture the structured information among the query and examples in the gallery.

%Inspired by previous ranking-motivated structured losses, we propose ranked list loss (RLL) to optimise the image retrieval results of each query directly. Given a query, we update its embedding by exploiting all positive and negative examples in the gallery.
%which is \textit{the most significant difference} from previous loss functions in DML.  
%
%The proposed RLL is illustrated in Figure~\ref{fig:ranked_list_loss}.
%
%
%As illustrated in Figure~\ref{fig:ranked_list_loss},

Secondly, we observe that the intraclass data distribution is not considered in the previous structured losses. All algorithms \cite{schroff2015facenet,song2016deep,sohn2016improved,song2017deep,movshovitz2017no,triantafillou2017few,goldberger2005neighbourhood} target to pull data points in the same class as close as possible. Consequently, these approaches try to \textit{shrink samples of the same class into one point} in the feature space and may easily drop their similarity structure. 
%In some classes, some examples are less similar than others, it is not proper to project every class to one point. 
To solve this, we propose to learn a hypersphere for each class in RLL. Specifically, instead of pulling intraclass examples as compact as possible, we only force the distance of a positive pair smaller than a threshold, which is the diameter of each class's hypersphere. In this case, RLL can explicitly preserve the intraclass similarity structure inside each class within the hypersphere.

%Given an anchor (query), the objective of RLL is to make the smallest distance of negative examples (the negative set with respect to the anchor) to be larger than the largest distance of positive examples (the positive set with respect to the anchor) by a margin. To the best of our knowledge, we are \textit{the first to group the positive and negative examples separately and force a margin between them} to improve the generalisation capability. 
%The comparison of our method with previous ranking-motivated losses is presented in Figure~\ref{fig:ranking_loss_comparison}.
%Among these methods, only the triplet loss (point-based) and our RLL (set-based triplet loss) force a margin between the negative and positive data points.

Empirically, the convergence rate of DML methods highly depends on the possibility of seeing non-trivial samples \cite{schroff2015facenet}.
Given a query (anchor), it is non-trivial to \textit{separate the positive and negative sets by a margin} when all data points are considered. 
As a result, only a few ranked lists are perfectly optimized as the model improves during training. 
%In contrast, for each epoch, 
Therefore, 
our method can take advantage of a maximum of elements with non-zero losses and release the potentials for the learning procedure.
%Therefore, our method can always see a large proportion of ranked lists with non-zero loss during training. 
%\note{The meaning is still not clear. A think this remark should be related to optimization during training and batch size. Because the number of non trivial elements (with non zero losses) is statistically low, and then we could accelerate the learning process if we are able to increase their quantities inside a mini-batch with a selective search, or if we increase the batch size.} 
The proposed RLL is illustrated in Figure~\ref{fig:ranked_list_loss}.

%\note{Critical Learning Period} 
Furthermore, a recent work \cite{achille2019critical} studied the existing of critical learning periods in artificial learning systems, which represents the time during which a temporary stimulus deficit may cause a permanent skill impairment in biological systems. 
Specifically, it shows the critical period in classification tasks using Fisher Information and Information Plasticity. Instead, we study and validate the critical learning period in deep metric learning via dynamic example weighting.  
Verifying the existing of critical learning periods is important, which can help us understand why network initialisation and warming-up training strategies have a large impact on the final performance. More importantly, it inspires us to be more careful and spend more effort on monitoring the early learning process of a deep network in the future work.   

%\note{Summarising Contributions}
In short, our contributions in this paper are listed as follows:
%\indent(1).
%\vspace{-7pt}
\begin{itemize}
	\item We propose a novel ranking-motivated structured loss, named Ranked List Loss (RLL), to learn discriminative embeddings using the recent setting of mini-batch few-shot retrieval.
	In contrast with previous ranking-motivated losses, we incorporate all non-trivial data points and exploit the structure among them. Novelly, we learn a hypersphere for each class to preserve intraclass data distribution instead of shrinking each class into one point in the embedding space. 
	%Given an query, only a proportion of non-trivial examples are considered in the previous ranking-motivated losses while RLL exploits all informative positive and negative data points in the gallery.\\
	%\indent
	%(2). The proposed RLL can be seen as an extension of the traditional point-based triplet loss into set-based version. Given an anchor, the objective of RLL is to separate the positive and negative sets in the gallery and force a margin between them. \\
	%\indent(2). 
	%\vspace{-7pt}
	
	\item 
	We propose two versions of RLL, i.e., the full version and a simplified version termed RLL-Simpler. 
	As indicated by its name, RLL-Simpler simplifies the full version and is more preferable for exploration in practice, because it has only two hyper-parameters.  
	
	\item Using RLL-Simpler, we obtain the state-of-the-art performance on two large datasets, i.e., SOP \cite{song2016deep} and In-shop Clothes \cite{liu2016deepfashion}.
	%Additionally, we present many other interesting results, which are of high practical value and can be open leads of future research. 
	%competitive performance compared with the state-of-the-art on popular benchmarks, including , CARS196 \cite{krause20133d} and CUB-200-2011 \cite{wah2011caltech}. 
	%
	%\item 
	Then, using the full version of RLL, we present comprehensive ablation studies for understanding the vital factors in deep metric learning. Specifically, the study on the network depth is of high practical value.

\end{itemize}

The rest of this paper is organised as follows: 
Section~\ref{section:preliminary} introduces some basic notations and preliminaries of deep metric learning, e.g., how a loss function is related to deep metric learning, prior work and practical strategies. 
We represent our method in section~\ref{section:methodology}. 
Specifically, in section~\ref{section:critical_learning_period}, we introduce a dynamic weighting scheme and use it to study the critical learning periods in the context of deep metric learning for the first time. 
In section~\ref{section:experiment}, we show extensive experiments to compare with related baselines and comprehensively study important components.
%Our implementation details of RLL are presented in section~\ref{section:implemtation_details}.  
The critical learning period of deep metric learning is studied in section \ref{section:critical_periods}. 
Finally, we make a summary of this work in section~\ref{section:conclusion}.

%%%%%%%%% Related Work & Prelimilaries
\section{Preliminaries and Related Work}
\label{section:preliminary}

\noindent{\textbf{Notations.}}
We use bold capital characters to denote matrices and set. 
Bold lower-case letters are used to represent vectors.  Scalar variables are denoted by non-bold letters. 
Let $\mathbf{X} = \{(\mathbf{x}_i, y_i)\}_{i=1}^{N}$ be the input data, where $(\mathbf{x}_i, y_i)$ indicates $i$-th image and its corresponding class label. The total number of classes is $C$, i.e., $y_i \in [1,2,...,C]$.    
The images from $c$-th class are represented as $\{\mathbf{x}_i^c\}_{i=1}^{N_c}$, where $N_c$ is the number of images in $c$-th class. 
For every $\mathbf{x}_i$, we use $y_i$ and superscript interchangeably to indicate its class label.

%%%%%%%%%%%%%%%%%%%%%%%%%%%%%%%%%%%%%%%%%%%%%%%%%%%%%
\begin{figure*}[t]
	%\vspace{0.8cm}
	\centering
	\includegraphics[width=\linewidth]{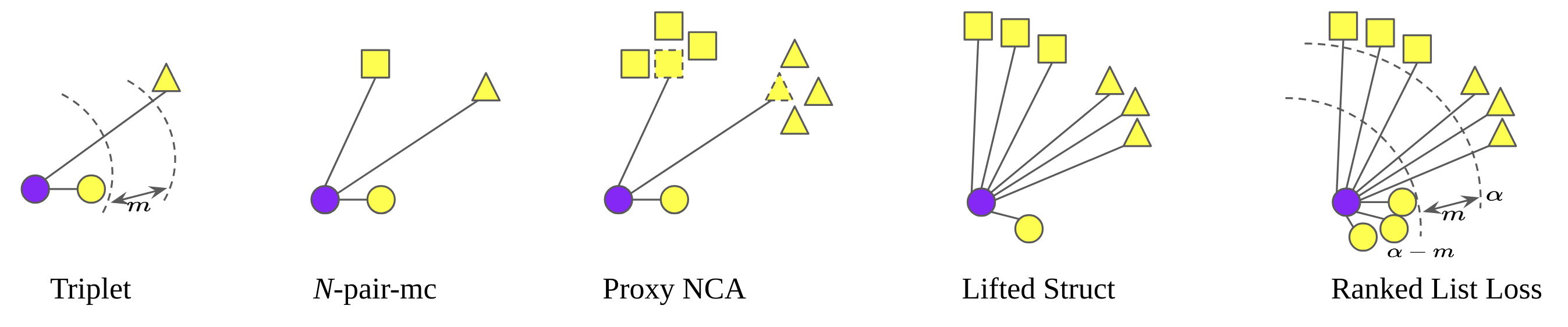}
	\vspace{-0.65cm}
	\caption{
		Illustration of different ranking-motivated structured losses. Different shapes (circle, triangle and square) represent different classes. For simplicity, only 3 classes are shown.
		The purple circle is an anchor (query).  
		In triplet~\cite{schroff2015facenet}, the anchor is compared with only one negative example and one positive example.  
		In $N$-pair-mc~\cite{sohn2016improved}, Proxy-NCA~\cite{movshovitz2017no} and Lifted Struct~\cite{song2016deep}, one positive example and multiple negative classes are incorporated. 
		$N$-pair-mc randomly selects one example per negative class.   
		Proxy NCA pushes the anchor away from negative proxies instead of negative examples. The proxy is class-level and can represent any instance in the corresponding class.
		Lifted Struct uses all examples from all negative classes. 
		On the contrary, our proposed Ranked List Loss not only exploits all negative examples, but also makes use of all positive ones. \textit{Best viewed in colour.}  
		%Besides, RLL forces a margin between the positive and negative set, thus being an extension of the point-based triplet loss.  
		%\textit{Best viewed in colour.}
	}
	\label{fig:ranking_loss_comparison}
	\vspace{-0.15cm}
\end{figure*}
%%%%%%%%%%%%%%%%%%%%%%%%%%%%%%%%%%%%%%%%%%%%%%%%%%%%

\subsection{Design of Loss Functions for Learning Discriminative Deep Representations}

In this subsection, we briefly introduce the relationship between design of loss functions for learning discriminative representations and deep distance metric learning. 

Generally, metric learning aims to learn a metric to measure the distance between two samples. 
For example, NCA \cite{goldberger2005neighbourhood} learns a linear transformation  from an input space $\mathbf{X}$ to a metric space $\mathbf{A}\mathbf{X}$. 
%i.e., $f(\mathbf{x})=\mathbf{A}\mathbf{x}$. 
In NCA, $d(\mathbf{x}_i, \mathbf{x}_j) = (\mathbf{A}\mathbf{x}_i-\mathbf{A}\mathbf{x}_j)^\top (\mathbf{A}\mathbf{x}_i-\mathbf{A}\mathbf{x}_j) =(\mathbf{x}_i-\mathbf{x}_j)^\top \mathbf{A}^\top \mathbf{A} (\mathbf{x}_i-\mathbf{x}_j)$. 
Here $\mathbf{A}^\top \mathbf{A}$ is a learned metric while $d(\mathbf{x}_i, \mathbf{x}_j)$ is the distance between $\mathbf{x}_i$ and $\mathbf{x}_j$ in the metric space.
However, \textbf{\textit{from the perspective of feature learning}}, metric learning learns an embedding function $f$, e.g., $f(\mathbf{x})=\mathbf{A}\mathbf{x}$. In the embedding (metric) space, the distance between every pair is computed by their Euclidean distance. A loss function, e.g., NCA for preserving neighbourhood structure, is needed to supervise the learning of an embedding function $f$. 
Therefore, we can see that the loss function defines how we compute the distance between two instances in the feature space. 
As a result, the design of a loss function is crucial in metric learning (discriminative embedding learning).

In the context of deep metric learning, a deep neural network is used as a non-linear encoding function. 
To empirically provide supervision information in the loss layer, there are three widely applied approaches: 
\begin{itemize}
	\item \textit{Learning to fulfil instance-to-class similarity relationship constraints}, i.e., pulling an example's feature towards its class centre while pushing it away from other class centres simultaneously. After encoding input data points, this approach is usually implemented with a fully connected layer for computing the dot product between an example's feature and all class centres, i.e., the weight vectors of the fully connected layer. The outputs are termed logits, which are normalised with a softmax transformation. Finally, the cross entropy is applied to ensure the instance-to-class similarity relationship constraint is met.  There are many variants of this approach, including 
	L2-Softmax \cite{ranjan2017l2}, Large-margin Softmax \cite{liu2016large}, Angular Softmax \cite{liu2017sphereface}, NormFace \cite{wang2017normface}, AM-Softmax \cite{wang2018additive}, CosFace \cite{wang2018cosface} and ArcFace \cite{deng2018arcface}.

	\item \textit{Learning according to the instance-to-proxy similarity relationship constraints.} 
	This idea is an interesting extension of instance-to-class similarity relationship modelling. 
	In this approach, classes are represented by proxies \cite{movshovitz2017no,qian2019softtriple}. 
	It is interesting in that more flexibility is given: (1) The number of proxies can be smaller than the number of training classes, in which case multiple classes are assigned to the same proxy. It is named fractional proxy assignment, i.e., one proxy represents multiple classes \cite{movshovitz2017no}; 
	(2) We can also represent one class using multiple proxies \cite{movshovitz2017no,qian2019softtriple};
	(3) When semantic class labels are available, we can apply static proxy assignment analogously to  instance-to-class modelling. 
	In cases where semantic labels are not given, we need to exploit dynamic proxy assignment according to the distances between a data point with all proxies \cite{movshovitz2017no}.  
	(4) When the number of proxies is the same as the number of classes and static proxy assignment is applied, this approach becomes equivalent to instance-to-class similarity relationship modelling.

	\item \textit{Leaning to meet instance-to-instance similarity relationship constraints.} Intrinsically, high-order similarity relationship, i.e., ranking motivated algorithms, are derived over pairwise relationship. In this context, we label that two data points are either similar or dissimilar, without the need to know how many training classes are given and which class each sample belongs to.  The proposed methods according to instance-to-instance similarity relationship include contrastive loss \cite{hadsell2006dimensionality,chopra2005learning}, triplet loss \cite{hoffer2015deep,schroff2015facenet}, Lifted Struct \cite{song2016deep}, N-pair-mc \cite{sohn2016improved}, Histogram loss \cite{ustinova2016learning}, Angular loss \cite{wang2017deep}, Sampling Matters \cite{wu2017sampling}, OSM and CAA \cite{wang2019deep}, Multi-similarity Weighting \cite{wang2019multi}, ICE \cite{wang2019instance} and so on. 
\end{itemize}

In this work, we focus on studying the second approach. Compared with instance-to-class or instance-to-proxy similarity relationship constraints, pairwise constraint is a good solution for the challenging extreme object recognition setting \cite{prabhu2014fastxml,yen2016pd}, in which there exist an enormous number of classes and only a few images per class. In our experiments, we evaluate our proposed method in such setting. Additionally, pairwise similarity constraint is more suitable for incremental (online) learning setting. When new training classes come, by instance-to-instance distance metric learning, we do not need to verify whether those new coming classes have occurred in the previous training stage. 
Beyond, knowledge transfer and re-learning is also more straightforward since we do not need to retrain a new fully connected layer.   
Theoretically, instance-to-instance similarity relationship modelling is more applicable for scaling up vision recognition. 

\subsection{Structured Losses}
\subsubsection{Ranking-Motivated Structured Losses}

\indent\emph{\textbf{Triplet Loss}}
\cite{weinberger2006distance,schroff2015facenet} aims to pull the anchor point closer to the positive point than to the negative point by a fixed margin $m$:
\begin{equation}
L(\mathbf{X};f) = \frac{1}{|\mathbf{\Gamma}|} \sum_{(i,j,k) \in \mathbf{\Gamma}} 
[
d_{ij}^2 + m - d_{ik}^2 
]_{+},
\end{equation}
where $\mathbf{\Gamma}$ is the set of triplets, $i,j$ and $k$ are the indexes of anchor, positive and negative points, respectively. $f$ is the embedding function, $d_{ij} = ||f(\mathbf{x}_i)-f( \mathbf{x}_j)||_2$ is the Euclidean distance. $[\cdot]_+$ is the hinge function.

%%%%%%%%%%%%%%%%%%%%%%%%%%%%%%%%%%%%%%%%%%%%%%
\emph{\textbf{\textit{N}-pair-mc}} \cite{sohn2016improved} exploits the structured relationship among multiple data points to learn the embedding function. Triplet loss pulls one positive point while pushing a negative one simultaneously. 
%Triplet loss suffers from slow convergence as a large number of triplets become trivial as the model improves. 
%As a result, triplet loss requires expensive sampling to provide informative triplets. 
To improve the triplet loss by interacting with more negative classes and examples, $N$-pair-mc aims to \textit{identify one positive example from $N-1$ negative examples of $N-1$ classes} (one negative example per class):
%\vspace{-10pt}
\begin{equation}
\begin{aligned}
L(\{(\mathbf{x}_i,\mathbf{x}_i^+)\}_{i=1}^N;f) = &  
\frac{1}{N} 
\sum_{i=1}^{N} log\{1+
\\
&
\sum_{j \neq i} \exp(\mathbf{f}_i^\top \mathbf{f}_j^{+} - \mathbf{f}_i^\top \mathbf{f}_i^+)
\}
\end{aligned},
\end{equation}
where $\mathbf{f}_i = f(\mathbf{x}_i)$ and $ \{(\mathbf{x}_i, \mathbf{x}_i^+)\}_{i=1}^N$ are $N$ pairs of examples from $N$ different classes, i.e., $y_i \neq y_j, \forall i \neq j$. Here, $\mathbf{x}_i$ and $\mathbf{x}_i^+$ are the query and the positive example respectively. $\{\mathbf{x}_j^+, j \neq i\}$ are the negative examples. 

%Compared with $N$-pair-mc, 
%which involves one positive example and one negative example per class
%our method \textit{ optimises the query by exploiting all positive examples and all negative examples from negative classes}. In addition, expensive hard negative `class' mining is required in $N$-pair-mc to provide non-trivial classes. 
%\emph{Angular Loss} \cite{wang2017deep}: \\

%%%%%%%%%%%%%%%%%%%%%%%%%%%%%%%%%%%%%%%%%%%%%%%
\emph{\textbf{Lifted Struct}} \cite{song2016deep} is proposed by Song \textit{et al.} to learn the embedding function by \textit{incorporating all negative examples}. 
%Given a positive pair $(\mathbf{x}_i, \mathbf{x}_j)$, 
The objective of Lifted Struct is to pull one positive pair ($\mathbf{x}_i^+$, $ \mathbf{x}_j^+$)
as close as possible and pushes all negative data points corresponding to $\mathbf{x}_i^+$ or $\mathbf{x}_j^+$ farther than a margin $\alpha$. Mathematically: 
\begin{equation}
\begin{aligned}
L(\mathbf{X};f) = &\frac{1}{2|\mathbf{P}|} \sum_{(i, j) \in \mathbf{P}} 
[
\{
d_{ij} + \log(
\sum_{(i, k) \in \mathbf{N}}\exp(\alpha-d_{ik})
\\
&
+\sum_{(j, l) \in \mathbf{N}}\exp(\alpha-d_{jl})
)
\}
]_{+}
\end{aligned},
\end{equation}
where $\mathbf{P}$ and $\mathbf{N}$ respectively represent the sets of positive pairs and negative pairs. 
Given the query $\mathbf{x}_i$, Lifted Struct intends to \textit{identify one positive example from all corresponding negative data points}. 

%Our method can be viewed as an extension of Lifted Struct. Given a query, \textit{we optimise the ranked list composed of all positive data points and all negative data points. }

\emph{\textbf{Proxy-NCA}} \cite{movshovitz2017no} is proposed to address the sampling problem using proxies.  
The proxy $\mathbf{W}$ is a small set of data points that represent training classes in the original data. The proxy for $\mathbf{u}$ is chosen by: 
\begin{equation}
p(\mathbf{u}) = \mathrm{arg min}_{\mathbf{w} \in \mathbf{W}} d(\mathbf{u},\mathbf{w}),
\end{equation}
$p(\mathbf{u})$ denotes the closest point to $\mathbf{u}$ from $\mathbf{W}$. The Proxy-NCA loss is the traditional NCA loss defined over proxies instead of the original data points:
\begin{equation}
L(\mathbf{a},\mathbf{u},\mathbf{Z})=-log(
\frac{\exp(-d(\mathbf{a},p(\mathbf{u}))))}
{\sum_{\mathbf{z} \in \mathbf{Z}} \exp(-d(\mathbf{a},p(\mathbf{z})))}
),
\end{equation}
where $\mathbf{Z}$ is the negative set, $p(\mathbf{u})$ and $p(\mathbf{z})$ are the proxies of positive and negative points, respectively. $\mathbf{a}$ is the anchor and $d(\cdot, \cdot)$ is the Euclidean distance between two points. 
With static proxy assignment, i.e., \emph{one proxy per class}, the performance is much better than dynamic proxy assignment. 
However, the proxies in the static proxy assignment are learned during training and similar to the class vectors of the fully connected layer in classification.
Recently, SoftTriple \cite{qian2019softtriple} further improves this proxy-based idea and achieves promising performance.  
The main drawback is that the scalability to extremely large datasets is not theoretically guaranteed. 
% which is the common trick of training a classifier, and using its penultimate layer’s output as the embedding.
%Therefore, \emph{Proxy-NCA does not preserve the scalability of DML} as the number of classes needs to be considered.

The proposed RLL is ranking-motivated structured loss, which avoids two limitations of traditional methods by incorporating all non-trivial data points and exploring intrinsic structured information among them. The illustration and comparison of different ranking-motivated losses and ours are presented in Figure~\ref{fig:ranking_loss_comparison}.

\subsubsection{Directly Learning to Retrieve}

Inspired by information retrieval, there are many methods proposed to maximise mean Average Precision over retrieved results of queries \cite{liu2009learning,mcfee2010metric,lim2014efficient}. We briefly present some recent representative work as follows. 

\emph{\textbf{Information Retrieval Lens}} \cite{triantafillou2017few} proposes a new form of a few-shot learning task, i.e., few-shot retrieval. Every batch is an independent task, composed of $n$ classes and $k$ images per class. Instead of splitting a batch into a support set (a training set) and a query set as done in Prototypical Networks \cite{snell2017prototypical},  Information Retrieval Lens \cite{triantafillou2017few} iteratively uses each data point in the batch as a `query' to rank the remaining ones. The optimisation objective of  \cite{triantafillou2017few} is to maximise the mean Average Precision (mAP) over those rankings. 

\emph{\textbf{FastAP}} \cite{cakir2019deep} approximately optimises the mean Average Precision via distance quantization. Optimising mAP directly is highly challenging, thus FastAP exploits quantization-based approximation to reduce the complexity and improve the efficiency. Therefore, FastAP is tailored for optimising mAP using stochastic gradient descent.    

\emph{\textbf{Prec@$K$}} \cite{lu2019sampling} means top-$K$ precision, or Recall@$K$ performance metric, which targets at optimising its top $K$ retrieved neighbours. To optimise Prec@$K$, it proposes to emphasise on misplaced images near the decision boundary, i.e., besides the $K$-th nearest neighbours. Concretely, those prioritised misplaced images are: (1) similar ones which do not belong to $K$ nearest neighbours, but are close to $K$-th nearest neighbour. The optimisation objective is to pull them into $K$ most nearest set; (2) dissimilar ones which are in the $K$ nearest neighbours set, and are close the $K$-th nearest neighbour. The optimisation target is to push them out of the $K$ nearest neighbours set. 

\textit{\textbf{MIHash}} \cite{cakir2019hashing} is a supervised hashing method for learning binary vector embeddings. For every query, MIHash minimises its neighbourhood ambiguity through an information-theoretic lens, where the separability between positive and negative points is measured by the mutual information between the distance distribution of positive examples and that of negative ones.  

In this paper, we also show that our proposed algorithm follows the setting of few-shot retrieval \cite{triantafillou2017few}. It optimises the distances of positive and negative pairs instead of average precision. 
Additionally, Prec@$K$ is partially motivated by our CVPR 2019 conference version \cite{wang2019ranked}.
{{FastAP}} \cite{cakir2019deep} is published in CVPR 2019, being concurrent with ours. The focus of MIHash is to learn a Hamming space for efficent retrieval in large databases.  

%\cite{cakir2019hashing}

\subsubsection{Clustering-Motivated Structured Losses}

\emph{\textbf{Struct Clust}} \cite{song2017deep} is recently proposed to learn the embedding function $f$ by optimising the clustering quality metric. The proposed structured loss function is defined as:    
\begin{equation}
L(\mathbf{X};f)=[
F(\mathbf{X}, \hat{\mathbf{y}};f) + \gamma \triangle(\mathbf{y}, \hat{\mathbf{y}}))
-F(\mathbf{X}, \mathbf{y};f)
]_{+},
\end{equation}
\begin{equation}
\triangle(\mathbf{y}, \hat{\mathbf{y}}) = 1 - \mathrm{NMI}(\mathbf{y}, \hat{\mathbf{y}}),
\end{equation}
where $\hat{\mathbf{y}}$ and $\mathbf{y}$ are the predicted and ground-truth clustering assignments respectively. $F$ measures the quality of the clustering on $\mathbf{X}$ with the label assignment and distance metric. 
$\mathrm{NMI}(\mathbf{y}, \hat{\mathbf{y}})$ is the normalised mutual information \cite{schutze2008introduction}. NMI is 1 if the predicted clustering assignment is as good as the ground-truth and 0 if it is the worst.  
$\hat{\mathbf{y}}$ is predicted based on the learned distance metric $f$ and Struct Clust \cite{song2017deep} aims to learn $f$ such that the $F$ of the ground-truth assignment is larger than any other predicted clustering assignment.  

However, this algorithm is NP-hard as we need to optimise both the clustering medoids and the distance metric simultaneously. As a result, the loss augmented inference and refinement are applied to select facilities (clustering medoids) based on the greedy algorithm \cite{mirzasoleiman2015lazier}. Large enough greedy search iterations are needed to find a local optimum, which might be costly. 

\emph{\textbf{Spectral Clust}} \cite{law2017deep} also aims to optimise the quality of the clustering.  
%Generally, it is difficult to optimise the clustering quality directly as shown in Struct Clust \cite{song2017deep}.
Spectral Clust relaxes the problem of clustering with Bregman divergences \cite{banerjee2005clustering} and computes the gradient in a closed-form, which reduces the algorithmic complexity of  existing iterative methods, e.g., Struct Clust \cite{song2017deep}. 
However, it is still non-trivial to learn deep models based on mini-batch implementation. Large batch size (i.e., 1260 = 18 classes $\times$ 70 samples per class) is required for the clustering in the mini-batch. As a result, Spectral Clust iteratively computes submatrices and concatenates them into a single matrix for computing the loss and gradient, which is computationally expensive. 

%\note{More methods introduction?}

Both ranking-motivated and clustering-motivated structured loss functions exploit the structured similarity information among multiple data points. However, in general, clustering-motivated losses are more difficult to optimise than ranking-motivated losses.

\subsection{Mining and Weighting Non-trivial Examples}

Example mining strategies are widely applied in existing methods \cite{lu2019sampling,schroff2015facenet,wang2015unsupervised,simo2015discriminative,huang2016local,yuan2017hard,shi2016embedding,cui2016fine,song2016deep,sohn2016improved,wang2019deep,wu2017sampling} to provide non-trivial examples for faster convergence and better performance. 
Mining strategies vary in different cases. 
For example, FaceNet \cite{schroff2015facenet} proposes to mine semi-hard negative samples.  
In $N$-pair-mc \cite{sohn2016improved}, hard negative class mining is proposed to provide informative negative examples. 
In Lifted Struct \cite{song2016deep}, harder negative examples are emphasized in a soft way. 
Given a query, Sampling Wisely \cite{lu2019sampling} proposes to select misplaced images near the decision boundary of its top-$K$ nearest neighbours.
Namely, Sampling Wisely \cite{lu2019sampling} prioritises data points near the top-$K$ decision boundary. 
In Divide \& Conquer \cite{sanakoyeu2019divide}, the embedding space is split into non-overlapping subspaces. Accordingly, the training data is split into clusters for training embedding functions in different subspaces. Consequently, two samples from the same subspace have smaller distances  than two from different clusters, which can be interpreted as a proxy to the mining of meaningful relationships. 
Stochastic class-based hard example mining \cite{suh2019stochastic} is proposed to mine hard examples effectively. 
%All the mined negative examples are treated equally in $N$-pair-mc. 
%
%In our work, we simply mine examples which have non-zero losses. 
%The naive example mining is applied in our method, i.e., 
%As the number of negative examples is much larger in each ranked list, we process negative and positive examples separately and applies independent normalisation \cite{wang2019deep}. \\

% \noindent
% \emph{\textbf{In summary}}, both ranking-motivated and clustering-motivated structured loss functions exploit the structured similarity information among multiple data points. Generally, clustering-motivated losses are more difficult to \note{organize} than ranking-motivated losses. Our proposed algorithm is ranking-motivated, thus \textit{being \note{easier} to \note{organize}}. 
%
%Among these methods, only the triplet loss (point-based) and our RLL (set-based triplet loss) force a margin between the negative and positive data points.
%Compared with existing ranking-motivated losses, \textit{for the first time we exploit the information from all informative examples}.  
%Empirically, we find this improves the performance a lot.
%

%%%%%%%%% Related Work Finished

\section{Methodology}
\label{section:methodology}
%%
%%No need to reformulate margin loss: combine with sampling
%%
Our objective is to learn a discriminative function $f$ such that similarity scores of positive pairs are higher than those of negative pairs in the feature space. There exist at least two samples in each class so that all data points can be optimised. Given a query from any class, the objective is to rank its matching samples in front of its retrieved result.

\subsection{Pairwise Constraint}

%Although many advanced loss functions have been studied in the previous work, e.g., the triplet loss \cite{weinberger2009distance} and structured losses \cite{song2016deep,song2017deep,sohn2016improved}, pairwise connection and similarity constraint~\cite{hadsell2006dimensionality,wu2017sampling} is an essential component of these loss functions. 
Inspired by the former work on pairwise similarity constraint~\cite{hadsell2006dimensionality,wu2017sampling}, we aim to pull positive examples closer than a predefined threshold (boundary). In addition, we intend to separate the positive and negative sets by a margin $m$. To achieve this, we choose the pairwise margin loss~\cite{wu2017sampling} as our basic pairwise constraint to construct the set-based similarity structure. 
Specifically, given an image $\mathbf{x}_i$, the learning objective is to push its negative points farther than a boundary $\alpha$ and pull its positive ones closer than another boundary $\alpha-m$. Thus $m$ becomes the margin between two boundaries.  
Mathematically,
\begin{equation}
\label{equation:margin_loss}
L_{\mathrm{m}}(\mathbf{x}_i,\mathbf{x}_j;f) = (1-y_{ij})[\alpha-d_{ij}]_{+} + y_{ij}[d_{ij}-(\alpha-m)]_+,
\end{equation}
where $y_{ij}=1$ if $y_{i}=y_{j}$, and $y_{ij}=0$ otherwise.
$d_{ij}=||f(\mathbf{x}_i)-f(\mathbf{x}_j)||_2$ is the Euclidean distance between two points. 
$[\cdot]_+$ is the hinge function.
%We will discuss this in section~\ref{section:RLL}.   

%Contrastive loss pushes the distances of all negative pairs beyond a margin $\alpha$ and pulls positive pairs as close as possible. 

%%%%%%%%%%%%%%%%%%%%%%%%%%%%%%%%%%%%%%%%%%%%%%%%%%%%%
\begin{figure*}[t]
	%\vspace{1.0cm}
	\centering
	\includegraphics[width=0.66\linewidth]{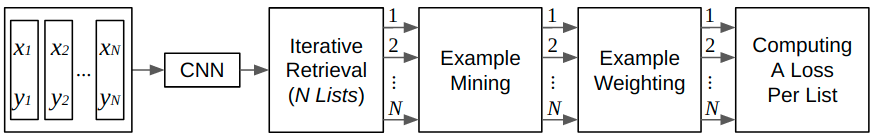}
	%\vspace{-0.22cm}
	\caption{
		The overall framework of our proposed ranked list loss. In one mini-batch, every image acts as a query iteratively and ranks other images according to the similarity scores. Then in every ranked list, we mine non-trivial data points and weight them based on their distances to the query. Finally, the ranked list loss is computed for every query.   
	}
	\label{fig:overall_framework}
	\vspace{-0.1cm}
\end{figure*}
%%%%%%%%%%%%%%%%%%%%%%%%%%%%%%%%%%%%%%%%%%%%%%%%%%%%

\subsection{Ranked List Loss}
\label{section:RLL}

%Compared with loss functions which only exploit the relations between doublets~\cite{hadsell2006dimensionality}, triplets~\cite{weinberger2009distance}, or quadruplet~\cite{huang2016local}, structured loss generally can learn more discriminative embeddings by using the structured information among multiple examples \cite{song2016deep,song2017deep,sohn2016improved}.
%
%However, in previous ranking-motivated loss functions, only a proportion of informative data points is incorporated to capture the structured information, which means many informative examples are wasted and fewer data points are included to build the structure. The illustration of different ranking-motivated structured losses is illustrated in Figure~\ref{fig:ranking_loss_comparison}.

%To address this, we propose \emph{ranked list loss} (RLL) to make use of all informative data points to build a more informative structure. As illustrated in Figure~\ref{fig:ranked_list_loss}, 
%given a query (anchor), the objective of RLL is to make any positive point closer to the query than any negative point in the gallery by a margin $m$. In addition, RLL pushes all negative examples farther than a predefined threshold $\alpha$. Consequently, the positive points are pulled closer than $\alpha-m$. Instead of pulling positive examples as close as possible in previous approaches~\cite{song2016deep,sohn2016improved,movshovitz2017no}, adding a margin between the positive and negative sets is more adapted and can improve the generalisation capability. 

In a mini-batch, when $\mathbf{x}_i^c$ is a query, we use it to retrieve the remaining data points, i.e., gallery according to their similarities to the query, which is illustrated in Figure~\ref{fig:ranked_list_loss}. 
In the retrieved list, there are $N_c - 1$ positive points in the positive set and $\sum_{k \neq c}N_k$ points in the negative set. The positive set with respect to the query $\mathbf{x}_i^c$ is denoted as $\mathbf{P}_{c,i}=\{\mathbf{x}_j^{c}|j \neq i\}, \text{ and } |\mathbf{P}_{c,i}|=N_c-1$. Similarly, we represent the negative set with respect to $\mathbf{x}_i^c$ as $\mathbf{N}_{c,i}=\{\mathbf{x}_j^{k}|k \neq c\}$, and $ |\mathbf{N}_{c,i}|=\sum_{k \neq c}N_k$.

%By incorporating contrastive loss as the basic pairwise module, our proposed RLL for optimising the ranked list of the query $\mathbf{x}_i^c$ is mathematically represented as: 

%\textbf{Non-trivial Sample Mining} 
%In Eq.~(\ref{equation:RLL_P}) and  Eq.~(\ref{equation:w_ij}), $d_{ij} \geq m$ and 
%$\alpha \geq d_{ij}$ represent mining non-trivial positive and negative samples respectively. By non-trivial samples, we mean positive and negative examples which have non-zero losses, i.e., violating the pairwise constraint with respect to the query. 
%We ignore the examples whose loss values are zeros. Since they have zero gradients, including them for training will `weaken' the few contributing examples during gradient fusion as the model improves.   
%Mining informative examples is very popular and more advanced mining strategies are adopted in previous methods \cite{schroff2015facenet,yuan2017hard,cui2016fine,song2016deep,sohn2016improved,wang2019deep} for faster convergence and better performance.

%%%Non-trivial Sample Mining%%%%%%

\subsubsection{Mining Informative Pairs} 
Mining informative examples is widely adopted \cite{schroff2015facenet,yuan2017hard,cui2016fine,song2016deep,sohn2016improved,wang2019deep,hermans2017defense}, because it can help to speed convergence and improve generalisation performance if properly designed. 
By informative examples, we mean non-trivial data points which have non-zero losses, i.e., violating the pairwise constraint with respect to a given query. 
Because trivial data pairs have zero gradients, including them for training can `weaken' the contribution of non-trivial examples during gradient fusion as the model improves \cite{hermans2017defense}. Therefore, we only train on non-trivial positive and negative examples. 

Concretely, for the query $\mathbf{x}_i^c$, the non-trivial positive set after mining is represented as $\mathbf{P}_{c,i}^*=\{\mathbf{x}_j^{c}|j \neq i, d_{ij} > (\alpha-m)\}$. Analogously, we denote the negative set after mining as $\mathbf{N}_{c,i}^*=\{\mathbf{x}_j^{k}|k \neq c, d_{ij} < \alpha \}$.
%%%Non-trivial Sample Mining%%%%%%

%%Loss-based Negative Examples Weightin%%%%

\subsubsection{Weighting Negative Pairs} 

For each query $\mathbf{x}_i^c$, there are a large number of non-trivial negative examples ($\mathbf{N}_{c,i}^*$) with different magnitude of losses. To make better use of them, we propose to weight the negative examples based on their loss values, i.e., how much each negative pair violates the constraint.   
Our weighting strategy can be simply represented as:
\begin{equation}
\label{equation:w_ij}
{w_{ij}}
= 
\exp(T_n \times (\alpha - d_{ij})),  ~~\mathbf{x}^k_j \in \mathbf{N}_{c,i}^*,
\end{equation}
%$\mathbf{x}^k_j$ is a non-trivial negative point of $\mathbf{x}^c_i$.
%Here, we use the square of Euclidean distance $d_{ij}=||f(\mathbf{x}^c_i)-f(\mathbf{x}^k_j)||^2_2$. 
%The reason is that the magnitude of partial derivatives of Euclidean distance can be regarded as weighting based on distances, i.e.,
%\begin{equation}
%|\frac{\partial ||f(\mathbf{x}^c_i)-f(\mathbf{x}^k_j)||^2_2}
%{\partial f(\mathbf{x}^k_j)}|
%=
%1.
%\end{equation}
%
%We notice that the gradient magnitude with respect to any embedding is always one in Eq.~(\ref{equation:margin_loss}). 
%%However, by using the square of Euclidean distance, the magnitude of partial derivatives is always one.  
%Mathematically,
%\begin{equation}
%||\frac{\partial L_{\mathrm{m}}(\mathbf{x}_i,\mathbf{x}_j;f)}
%{\partial f(\mathbf{x}_j)}||_2
%=||\frac{f(\mathbf{x}_i)-f(\mathbf{x}_j)}
%{||f(\mathbf{x}_i)-f(\mathbf{x}_j)||_2}||_2
%=1
%.
%\end{equation}
%Consequently, the gradient magnitude of any embedding is only determined by our weighting strategy $w_{ij}$.  In this case, it is also convenient to evaluate its influence, which is studied in section~\ref{section:ablation}.
%
%
%In Eq.~(\ref{equation:w_ij})
where $T_n$ is the temperature parameter which controls the degree (slope) of weighting negative examples. If $T_n=0$, we treat all non-trivial negative examples equally. If $T_n=+\infty$, it becomes the hardest negative example mining because weights are normalised by their sum.  
%%Loss-based Negative Examples Weightin%%%%

\subsubsection{Weighting Positive Pairs} 

Usually, in the settings of deep metric learning, given a query, there are quite a few matching positives in the search space.
However, when multiple positive data points exist, we can also weight those positive ones according to their loss values to make better use of them.    
Analogous to negative examples weighting, the weighting strategy of positive data pairs can be denoted as follows:
\begin{equation}
	\label{equation:pw_ij}
	{w_{ij}}
	= 
	\exp(T_p  \times (d_{ij} - (\alpha-m) )), ~~ \mathbf{x}^c_j \in \mathbf{P}_{c,i}^*,
\end{equation}
%
%
%In Eq.~(\ref{equation:pw_ij})$, 
where $T_p$ is the temperature parameter which controls the degree (slope) of weighting positive examples. If $T_p=0$, it treats all non-trivial positive data pairs equally. 
If $T_p > 0$, positive pairs with larger distances are emphasised.   
On the contrary, if $T_p < 0$, closer positive pairs are assigned with higher weights, which is widely used to preserve local similarity manifold structure when many positive data points exist \cite{huang2016local,cui2016fine,wang2019deep}.   
The absolute value of $T_p$ determines the differentiation degree over positive data pairs. 

%\textbf{Independent Normalisation}
%For each query $\mathbf{x}_i^c$, RLL optimises the positive list $L_{N}(\mathbf{x}_i^c;f)$ and  the negative list $L_{P}(\mathbf{x}_i^c;f)$ separately, denoted by Eq.~(\ref{equation:RLL_P}) and Eq.~(\ref{equation:RLL_N}) respectively. 
%
%This is because the number of positive images and negative images is imbalanced and the distance distribution of positive pairs and negative pairs is different  \cite{wang2019deep}.
%
%Although positive examples and negative examples are processed separately, they are connected by the distance constraint. RLL pulls all positive examples closer than $\alpha - m$ in Eq.~(\ref{equation:RLL_P}) while pushes all negative examples farther than $\alpha$ in Eq.~(\ref{equation:RLL_N}). As a result, RLL splits the positive set and negative set by a margin $m$, which is a set-based triplet loss and an extension of the point-based triplet loss \cite{schroff2015facenet}.  
%The illustration of RLL is shown in Figure~\ref{fig:ranked_list_loss}. 
%The difference is that we cut off easy positive points ($d_ij < \alpha-m$) and trivial negative examples (larger than )

%\subsection{Algorithmic Implementation Details}
%\label{section:Algorithmic_Implementation_Details}
%\subsubsection{All You Need Is Example Weighting}

\subsubsection{Overall Optimisation Objective}

%%%%Optimisation Objective%%%%%%%

In order to pull all non-trivial positive points in $\mathbf{P}^*_{c,i}$ together and learn a class hypersphere, we minimise:
%\begin{equation}
%\label{equation:RLL_P}
%L_\mathrm{P}(\mathbf{x}_i^c;f) 
%= 
%\frac{1}
%{|\mathbf{P}^*_{c,i}|}
%%
%\sum_{\mathbf{x}_j^{c} \in \mathbf{P}^*_{c,i}}
%L_{\mathrm{m}}(\mathbf{x}_i^c,\mathbf{x}_j^c;f).
%\end{equation}
\begin{equation}
\label{equation:RLL_P}
L_\mathrm{P}
(
\mathbf{x}_i^c
%,\{\mathbf{x}_j^k\}_{k \neq c}
;f
) 
= 
%\frac{1}{C}
\sum_{\mathbf{x}_j^c \in |\mathbf{P}^*_{c,i}|}
\frac{w_{ij}}{\sum_{\mathbf{x}_j^c \in |\mathbf{P}^*_{c,i}|}
	w_{ij}}
L_{\mathrm{m}}(\mathbf{x}_i^c,\mathbf{x}_j^c;f).
\end{equation}
%We do not weight positive points because there exist only a few positive examples. 
Meanwhile, to push the informative negative points in $\mathbf{N}^*_{c,i}$ beyond the boundary $\alpha$, we minimise:
\begin{equation}
\label{equation:RLL_N}
L_\mathrm{N}
(
\mathbf{x}_i^c
%,\{\mathbf{x}_j^k\}_{k \neq c}
;f
) 
= 
%\frac{1}{C}
\sum_{\mathbf{x}_j^k \in |\mathbf{N}^*_{c,i}|}
\frac{w_{ij}}{\sum_{\mathbf{x}_j^k \in |\mathbf{N}^*_{c,i}|}
	w_{ij}}
L_{\mathrm{m}}(\mathbf{x}_i^c,\mathbf{x}_j^k;f).
\end{equation}
%
%In RLL, we treat the two minimisation objectives equally and optimise them jointly: 
In RLL, we optimise the two minimisation objectives jointly: 
\begin{equation}
\label{equation:RLL_x_i}
L_\mathrm{RLL}(\mathbf{x}_i^c;f) 
= 
%\lambda_p
(1-\lambda)L_\mathrm{P}(\mathbf{x}_i^c;f) + 
\lambda
L_\mathrm{N}(\mathbf{x}_i^c;f),
\end{equation}
where $\lambda$ controls the balance between positive and negative sets. 
We treat the two objectives equally and fix $\lambda=0.5$ without tuning in all our experiments. In this case, the positive and negative sets can contribute equally. Consequently, the sample imbalance problem, i.e., the majority are negative examples, is addressed well. 
%This  works well in our practice.  

We remark that our proposed RLL is an extension of traditional point-based triplet loss. Given an anchor, RLL separates the positive and negative sets with a margin between them. 
Note that in the optimisation of every retrieved list, we exploit {independent normalisation} \cite{wang2019deep} to address the imbalanced number of positive and negative examples, which is {not considered in the previous ranking-motivated losses}.  
In the ranked list of $\mathbf{x}_i^c$, we regard the features of other examples as constants. Therefore, only $f(\mathbf{x}_i^c)$ is updated based on the influence of weighted combination of other data points, which makes the learning process more stable.

%%%%Optimisation Objective%%%%%%%

\subsection{Hypersphere Regularisation by Distance Thresholds}
\begin{figure}[t]
	%\vspace{0.2cm}
	\centering
	\includegraphics[width=\linewidth]{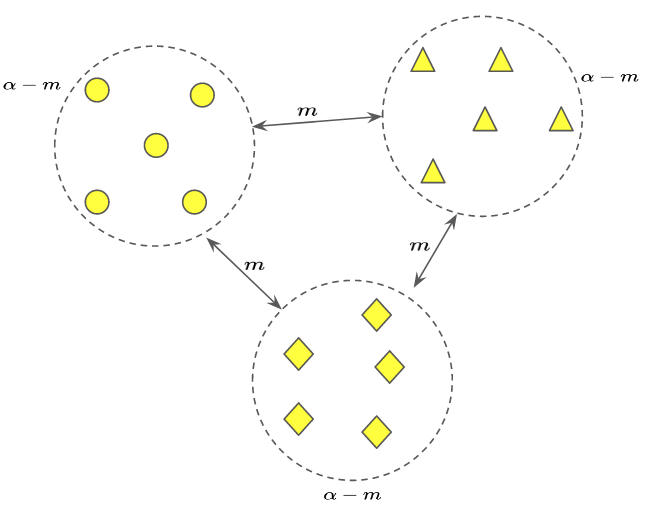}
	\vspace{-0.45cm}
	\caption{
		The optimisation objectives with hypersphere regularisation. 
		Different shapes represent different classes. For simplicity, we only show three classes while many classes exist in practice. 
		$\alpha-m$ denotes the diameter of each class hypersphere. Therefore, the distance between any two positive examples is optimised to be no greater than $\alpha-m$.  
		In addition, the distance between any two hypersphere boundaries is no less than $m$.   
	}
	\label{fig:hyperball}
	\vspace{-0.20cm}
\end{figure}

For each query $\mathbf{x}_i^c$, we propose to make it closer to its positive set $\mathbf{P}_{c,i}$ than to its negative set $\mathbf{N}_{c,i}$ by a margin $m$. At the same time, we force all negative examples to be farther than a boundary $\alpha$. Consequently, we pull all samples from the same class into a hypersphere. The diameter of each class hypersphere is $\alpha-m$. 
According to our extensive ablation studies on both large and small datasets, we find that it is an improper practice to pull positive data pairs as close as possible in the literature. Intuitively, we should not try to suppress all intraclass variances.  
By pulling all intraclass examples into one hypersphere instead of one point, better generalisation performance can be obtained. Therefore, we term it hypersphere regularisation. 
The optimisation objectives with hypersphere regularisation is illustrated in Figure~\ref{fig:hyperball}.

%%%%%%%%%%%%%%%%%%%%%%%%%%%%%%%%%%%%%%%%%%%%%%%%%%%%%
%%%%%%%%%%%%%%%%%%%%%%%%%%%%%%%%%%%%%%%%%%%%%%%%%%%%

%\subsection{Refine Example Weighting}
%In our proposed RLL, there are 

\subsection{RLL-Simpler}

We have introduced the full version of RLL, which includes two distance hyper-parameters $\alpha$ and $m$ for defining the optimisation objectives, and two scaling parameters $T_n$ and $T_p$ for weighting negative and positive data points, respectively.  
In addition, we propose a simpler version which has only two hyper-parameters, thus being much easier to apply in practice. 

Firstly, we can reformulate the optimisation objectives to use only one distance threshold. Following the widely used practice \cite{wang2017normface,song2017deep,law2017deep,movshovitz2017no}, we apply an $L_2$ normalisation layer after the final fully connected layer. As a result, the Euclidean distance any two data points ranges from 0 to 2.   
Recognising a data pair as similar or dissimilar can be treated as a binary classification problem. Therefore, intuitively, we can use the hyperplane Euclidean distance $=1$ as the decision boundary.  
Following this intuition, we change the setting of $\alpha$ and $m$ in RLL as follows: 
\begin{itemize}
	\label{equation:RLL_Simple_alpha}
	\item $\text{Smallest distance between negative points: }
	\alpha = 1 + \frac{m}{2};$
	\item $\text{Largest distance between positive points: } \alpha-m = 1-\frac{m}{2};$
	\item $\text{Margin between positive and negative set: } m.$
\end{itemize}
%\begin{equation}
%%\label{equation:RLL_Simple_alpha}
%\begin{aligned}
%&\text{Smallest distance between negative points: }
%\alpha = 1 + \frac{m}{2};\\
%&
%\text{Largest distance between positive points: } \alpha-m = 1-\frac{m}{2};
%\\
%& \text{Margin between positive and negative set: } m. 
%\end{aligned}
%\end{equation}
Namely, in RLL-Simpler, we aim to push the distances of negative pairs to be larger than $1 + \frac{m}{2}$ while pull those of positive ones to be smaller than $1 - \frac{m}{2}$. 

Secondly, training RLL on a mini-batch follows the few-shot retrieval setting, i.e., $N_c$-shot $C$-way setting. Generally, $C$ is much larger than $N_c$, e.g., $C=60 ~\text{and } N_c =3 , \forall c$,   therefore we have much more negative points than positive ones in the retrieved list. Consequently, by default, we set $T_p = 0$ without weighting positive data points in practice.   

In summary, in RLL-Simpler, we have only two active hyper-parameters: $m$ for defining the margin between positive and negative pairs, and $T_n$ for weighting negative examples of a query.
It is a simplified version of the full version by setting $T_p = 0$, and $\alpha = 1 + \frac{m}{2}$. 
With RLL-Simpler, our objective is to demonstrate that our proposal is able to achieve competitive performance without the need to manually optimise many hyper-parameters. 
%with the state-of-the-art, 
In real-world applications, we recommend first applying RLL-Simpler, and then adopting the full version to further push the performance if needed.

\subsection{Learning Deep Models with RLL}
\label{section:deep_model}
To learn deep models, we implement our RLL based on mini-batches and stochastic gradient descent. Each mini-batch is a randomly sampled subset of the whole training classes, 
which can be regarded as \textit{a mini ranking problem with a much smaller gallery}, i.e., searching the matching examples from a smaller number of classes. 

A mini-batch is composed of $C$ classes and $N_c$ images per class ($N_c$-shot $C$-way setting). Therefore, each mini-batch can be also treated as a few-shot retrieval task \cite{triantafillou2017few}.    
Every image $\mathbf{x}_i^c$ in the mini-batch acts as 
the query (anchor) iteratively and the other images serve as the gallery. The RLL on a mini-batch is represented as: 

\begin{equation}
\label{equation:RLL_mini_batch}
L_\mathrm{RLL}(\mathbf{X};f) 
= 
\frac{1}{N}
\sum_{\forall c, \forall i}
L_\mathrm{RLL}
(
\mathbf{x}_i^c
;f
),  
\end{equation}
where {$N=\sum_{c}N_c$ is the batch size}. The learning of the deep embedding function $f$ based on RLL is illustrated in Algorithm~\ref{algorithm:RLL}. The overall pipeline is shown in Figure~\ref{fig:overall_framework}.

\begin{algorithm}[t]
	\begin{algorithmic}[1]
		\State \textbf{Mini-Batch Settings}: The batch size $N$, the number of classes $C$, the number of images per class $N_c$. 
		\State \textbf{Parameters}: The distance constraint $\alpha$ on negative points, the margin between positive and negative examples $m$, the weighting temperature $T_n, T_p$.   
		\State \textbf{Input}: $\mathbf{X} = \{(\mathbf{x}_i, y_i)\}_{i=1}^{N} = \{\{\mathbf{x}_i^c\}_{i=1}^{N_c}\}_{c=1}^C $, the embedding function $f$, the learning rate $\beta$.  
		\State \textbf{Output}: Updated $f$.
		
		\State \textbf{Step 1}: Feedforward all images $\{\mathbf{x}_i\}_{i=1}^{N}$ into $f$ to obtain the images' embeddings $\{f(\mathbf{x}_i)\}_{i=1}^{N}$. 
		
		\State \textbf{Step 2}: Iterative retrieval and loss computation. 
		\Indent
		\ForAll{$f(\mathbf{x}_i^c) \in 
			\{\{f(\mathbf{x}_i^c)\}_{i=1}^{N_c}\}_{c=1}^C$}
		
		\State Mine non-trivial positive set $\mathbf{P}^*_{c,i}$.
		
		\State Mine non-trivial negative set $\mathbf{N}^*_{c,i}$.  
		
		\State Weight negative examples using Eq.~(\ref{equation:w_ij}).  
		
		\State Weight positive examples using Eq.~(\ref{equation:pw_ij}).  
		
		\State Compute $L_\mathrm{P}(\mathbf{x}_i^c;f)$ using Eq.~(\ref{equation:RLL_P}).
		%\Indent
		%\ForAll{$f(\mathbf{x}_j^c) \in \{f(\mathbf{x}_i^c)\}_{i=1}^{N_c}\}$}
		%\EndFor
		%\EndIndent
		
		\State Compute $L_\mathrm{N}(\mathbf{x}_i^c;f)$ using Eq.~(\ref{equation:RLL_N}).
		
		\State Compute $L_\mathrm{RLL}(\mathbf{x}_i^c;f)$ using Eq.~(\ref{equation:RLL_x_i}).	
		\EndFor
		\State Compute $L_\mathrm{RLL}(\mathbf{X};f)$ using Eq.~(\ref{equation:RLL_mini_batch}).
		\EndIndent
		\State \textbf{Step 3}: Gradient computation and back-propagation to update the parameters of $f$: 
		\Indent
		\State $\nabla_f=\partial L_\mathrm{RLL}(\mathbf{X};f)/\partial f$; 
		\State $f=f- \beta \cdot \nabla_f$.
		\EndIndent
		
	\end{algorithmic}
	\caption{Ranked List Loss on one mini-batch. %(RLL-Simpler has only two active hyper-parameters. Its algorithm's pipeline is the same.) 
	}
	\label{algorithm:RLL}
\end{algorithm}

\subsection{Computational Complexity} 
As illustrated in Algorithm~\ref{algorithm:RLL}, our proposed method does not require the input data to be prepared in any rigid format, e.g., triplets, n-pair tuplets. Instead, it takes random input images with multiclass labels.  We conduct online iterative ranking and loss computation (step 2 in Algorithm~\ref{algorithm:RLL}) after obtaining images' embeddings (step 1 in Algorithm~\ref{algorithm:RLL}). Therefore, the computational complexity of RLL is $O(N^2)$, which is the same as existing ranking-motivated structured loss functions \cite{song2016deep,sohn2016improved,movshovitz2017no}. 
%Specifically, in Triplet Semihard \cite{schroff2015facenet} $O(N^3)$,  in Lifted Struct~\cite{song2016deep} $O(N^2)$, in $N$-pair-mc~\cite{sohn2016improved}, $O(N^2)$, 

\subsection{Exploring The Critical Learning Periods of Deep Metric Learning}
\label{section:critical_learning_period}

In biological systems, critical period represents the time windows of early post-natal development during which a temporary stimulus deficit may lead to a skill impairment permanently \cite{olson1980profile,mitchell1988extent, giffin1978rate, konishi1985birdsong, wiesel1963effects, kandel2013principles}.                          
Critical periods do not exist only in biological systems, but also in artificial learning systems \cite{achille2019critical}. 
Recently, a study on the critical learning periods \cite{hensch2004critical} of deep neural networks is quite interesting and has got much attention \cite{achille2019critical}.
As a result of fundamental constrains coming from learning dynamics and information processing, \cite{achille2019critical} finds that the critical early transient determines the final optimised solution. Specifically, no matter how much additional training is used, a temporary stimulus deficit during the critical early learning period cannot be overcome later.   

The early learning phase of artificial deep neural networks is under-scrutinised compared to the network's behaviours around convergence and the asymptotic properties of the optimisation process. 
However, it plays a key role. In \cite{achille2019critical}, Fisher Information and Information Plasticity have been exploited to study the early learning phase. In our work, we validate and study this critical early learning phase in deep metric learning from a novel perspective, i.e., dynamic example weighting. 
We represent our design choices in this section and display the results in Section~\ref{section:critical_periods}. 

Concretely, during training, we revise Eq.~(\ref{equation:w_ij}) to its dynamic variant:  
\begin{equation}
\label{equation:modified_Tn}
{T_n}
= 
T_1 - cur\_iter * \frac{T_1-T_2}{max\_iter},
\end{equation}
\begin{equation}
\label{equation:dynamic_w_ij}
{w_{ij}}
= 
\exp( T_n \times (\alpha - d_{ij})),  \mathbf{x}^k_j \in \mathbf{N}_{c,i}^*,
\end{equation}
%To be more specific, we modify $T_n$ as: 
where $cur\_iter < max\_iter$, 
$max\_iter$ denotes the total number of training iterations while $cur\_iter$ is the performed number of iterations till now. 

Therefore, instead of fixing the scaling parameter of the weighting scheme, we can study and analyse the effect of dynamic weighting scheme on the optimisation results. 
Two cases we aim to explore are introduced in detail as follows. 
Their empirical results are presented and discussed in section~\ref{section:critical_periods}. 

\begin{itemize}
	\item {Study of The Early Learning Phase}. To validate the critical early learning phase in deep metric learning, we fix $T_2$ and present the results of different $T_1$. 
	In this case, the early learning phase changes along with $T_1$. Intuitively, if the early learning phase is critical, then the test performance will be sensitive to $T_1$.
	\item {Exploration of The Later Learning Phase}. On the contrary, we fix $T_1$ and change $T_2$ to study the effect of the later learning phase on the final optimisation solution. 
	Similarly, if the test performance is sensitive to $T_2$, then the later learning phase is crucial, and non-important otherwise. 
	
\end{itemize}

%%%%%%%%% 

\section{Experiments}
\label{section:experiment}

\subsection{Experimental Details}
\label{section:implemtation_details}

Some recent papers \cite{musgrave2020metric,fehervari2019unbiased} have raised the concerns about the fairness of comparing different DML methods. For example, Lifted Struct~\cite{song2016deep} reports that the embedding dimension does not play a crucial role. 
Accordingly, thereafter, the embedding size is different in some papers  \cite{song2016deep,song2017deep,law2017deep,sohn2016improved,ge2018deep}.
However, it is recently noticed that the embedding size has a huge impact on the performance \cite{musgrave2020metric,fehervari2019unbiased}.   
In this work, to make sure that our comparison is as fair as possible, we present all our implementation details as follows:  

\noindent
\subsubsection{Datasets} 
We conduct experiments on four popular benchmarks: 
(1) \emph{SOP} \cite{song2016deep} contains 120,053 images of 22,634 online products sold on eBay.com. 59,551 images of 11,318 categories and 60,502 images of 11,316 categories are used for training and testing respectively. The train/test split and evaluation protocol are the same as \cite{song2016deep}.
(2) \emph{In-shop Clothes} \cite{liu2016deepfashion} contains 7,982 classes and 52,712 images in total. It is split into a training set and a testing set. The training data contains 25,882 images of 3,997 classes. The testing set includes 3, 985 classes, and there are 14,218 query images and 12,612 in the gallery for search. 
(3) \emph{CUB-200-2011} \cite{krause20133d} has 11,788 images of 200 bird species. 5,864 images of the first 100 classes are used for training and 5,924 images of the other 100 classes for testing.
(4) \emph{CARS196} \cite{wah2011caltech} contains 16,185 images of 196 car models. We use the first 98 classes (8,054 images) for training and the remaining 98 classes (8,131 images) for testing. 
On all datasets, our method is evaluated on the original images, i.e., without using the bounding box information.   

%\noindent
\subsubsection{Data Augmentation} 
%In some papers, the data augmentation used in their released code is inconsistent with what presented in the paper. For example, in Multi-Simi \cite{wang2019multi}, it is written ``All the input images were cropped to 224 $\times$ 224. For data augmentation, we used random crop with random horizontal mirroring for training". 
%However, in fact, their implementation is: (1) warp the original image to a size of $256\times 256$; 
%(2) crop a random size (default: 0.16 to 1.0 of the resized image)\footnote{\url{https://github.com/MalongTech/research-ms-loss/blob/master/ret_benchmark/config/defaults.py\#L72} } with a random aspect ratio (default: 3/4 to 4/3)\footnote{\url{https://pytorch.org/docs/stable/_modules/torchvision/transforms/transforms.html\#RandomResizedCrop} }; 
%(3) This crop is finally resized to $227 \times 227$\footnote{\url{https://github.com/MalongTech/research-ms-loss/blob/master/ret_benchmark/config/defaults.py\#L73}} and horizontally flipped with a probability of 0.5.
%\url{https://pytorch.org/docs/stable/_modules/torchvision/transforms/transforms.html#RandomResizedCrop}
%\url{https://discuss.pytorch.org/t/is-transforms-randomresizedcrop-used-for-data-augmentation/16716}
%
%In our conference version, we simply crop a random size of $227\times 227$ from the resized image ($256\times 256$). In this version, whenever a fair comparison is needed, we follow their implementation \cite{wang2019multi}.
For fair comparisons, we follow the practice of \cite{wang2019multi} for data augmentation. Concretely, we (1) warp the original image to a size of $256\times 256$; 
(2) crop a random size (default: 0.5 to 1.0 of the resized image)\footnote{We change the random crop size from the range [0.16, 1.0] used in \cite{wang2019multi} to [0.5, 1] as we do not want a crop containing no object of interest at all.}
%\footnote{\url{https://github.com/MalongTech/research-ms-loss/blob/master/ret_benchmark/config/defaults.py\#L72} }
with a random aspect ratio (default: 3/4 to 4/3);
%\footnote{\url{https://pytorch.org/docs/stable/_modules/torchvision/transforms/transforms.html\#RandomResizedCrop} }; 
(3) resize the crop to $224 \times 224$
%\footnote{\url{https://github.com/MalongTech/research-ms-loss/blob/master/ret_benchmark/config/defaults.py\#L73}} 
and horizontally flip it with a probability of 0.5.
%\url{https://pytorch.org/docs/stable/_modules/torchvision/transforms/transforms.html#RandomResizedCrop}
%\url{https://discuss.pytorch.org/t/is-transforms-randomresizedcrop-used-for-data-augmentation/16716}
%We only change the random crop size from the range [0.16, 1.0] used in \cite{wang2019multi} to [0.5, 1] because we do not want a crop containing no object of interest at all.  
%we present an ablation study on data augmentation, specifically, the random scale range. We conjecture that  0.16 is too small, which leads to semantic noise during training.  
At testing stage, we only use a single centre crop without mirroring. 
We remark that the performance on large datasets (i.e., SOP and In-shop Clothes) changes slightly when the data augmentation is different. While small datasets CUB-200-2011 and CARS-196 are very sensitive to small changes. 
%Specifically, the input images are first resized to $256 \times 256$ and then cropped at $227 \times 227$.
%During training, we use random crop and random horizontal mirroring for data augmentation. 
%As reported in Lifted Struct~\cite{song2016deep}, the embedding dimension does not play a crucial role. 
%Therefore, 
%We set the embedding size to 512 on all datasets following the setting in \cite{law2017deep,sohn2016improved}.  

%As done in \cite{song2017deep,movshovitz2017no}, the features are $L_2$ normalised before computing their distance during training and testing. 
%

\subsubsection{Backbones and Initialisation Details}
 
Except for showing the results of ResNet-50 \cite{he2016deep} in Table~\ref{table:SOTA_SOP_RESNET}, 
we use GoogLeNet V2 \cite{ioffe2015batch} as our backbone network for all experiments, and compare with prior results using the same backbone as well. 
%for fair comparison with \cite{song2017deep,movshovitz2017no,law2017deep}. 
Additionally, in the original net, there are three fully connected layers of different depth. We refer them based on their relative locations as follows: L for the low-level layer (inception-3c/output), M for the mid-level layer (inception-4e/output) and H for the high-level layer (inception-5b/output). 
By exploiting them, we study the representations of different depth, which is valuable for deploying a model in practice. 
Following \cite{song2017deep,movshovitz2017no,law2017deep},  %\cite{song2017deep,movshovitz2017no,law2017deep} 
the pretrained model on ImageNet \cite{russakovsky2015imagenet} is used for initialisation in our experiments. Three original 1000-neuron fully connected layers followed by the softmax layer and cross-entropy loss are changed to three new fully connected layers followed by an $L_2$ normalisation layer and our proposed ranked list loss.  
The new layers are randomly initialised and optimised with 10 times larger learning rate than the others for faster convergence.  

For completeness, we also experiment with ResNet-50 backbone, which is also initialised by a pre-trained model on ImageNet. Moreover, ResNet-50 has single embedding layer in the head, which is randomly initialised and trained using 10 times larger learning rate than the other layers in the backbone.

%\subsubsection{Freezing BN Layers on CUB-200-2011 and CARS-196}

It is well accepted in \cite{wang2019multi,qian2019softtriple,musgrave2020metric} that CUB-200-2011 and CARS-196 are sensitive to small changes of training details. Therefore, following the concurrent work Multi-Simi \cite{wang2019multi}, and the following work SoftTriple \cite{qian2019softtriple} and Reality Check \cite{musgrave2020metric}, we freeze the BatchNorm layers during training to fairly compare with their reported results on CUB-200-2011 and CARS-196.

%\noindent
\subsubsection{Embedding and The Embedding Size} 
For GoogLeNet V2, 
we use RLL-H to denote the single high-level embedding. RLL-M and RLL-L are analogous.   
For an exactly fair comparison with other baselines, we look at the results of RLL-H. 
%
%Recently, it is noticed that the performance is sensitive to the embedding size in DML. 
Generally, there are two ways to increase the embedding size: (1) increasing the size of a single embedding;
(2) concatenating multiple embeddings. 
Empirically, we find that our method is less sensitive to the increasing size of a single embedding, while performs much better when concatenating multiple embeddings.

On all datasets, we follow their corresponding most common setting and set the embedding size accordingly to fairly compare with as many existing approaches as possible. 
Finally, we study the impact of the embedding size on the largest dataset SOP in the section~\ref{ablation_study:embedding_size}.
% so that we do not have to train multiple models.   

%\noindent
%\subsubsection{} 
%We study the impact of the embedding %size on the largest dataset SOP. 
%Thereafter, 

%\noindent
\subsubsection{Performance Metric}

Following \cite{song2016deep}, we report the image retrieval performance measured by Recall@\textit{K}. 
%and the image clustering quality and NMI~\cite{schutze2008introduction}, respectively. 
We do not report the image clustering quality NMI~\cite{schutze2008introduction}, because NMI is not a proper metric for fine-grained object recognition with tremendous classes and only several images per class. Similar idea is presented recently in very detail in \cite{musgrave2020metric}. 
%Therefore, we only report NMI on small datasets. 

Indeed, all reported results are validation performance. Theoretically, the validation data follows the same distribution as the test data, and the test data is always unknown before the deployment for practical use. Therefore, it is reasonable to report the validation performance as long as the validation data is only used for model selection, without being used for training. 
%For each query, the Recall@\textit{K} is 1 if any positive image appears in the top \textit{K} positions of the ranked list of the gallery. 
%\textbf{EK: Add definition of NMI or remove the definition of Recall@k}% While testing, every image serves as the probe iteratively and the average Recall@\textit{K}  is reported. 

%\subsection{Optimisation and Evaluation}
%\noindent
\subsubsection{Training and Optimisation Settings.} 
%We run our experiments on a single Tesla V100 GPU with 32 GB RAM. 
The standard stochastic gradient descent (SGD) optimiser is used with a momentum of 0.9, a weight decay rate of $2e^{-5}$. We set the base learning rate to $1e^{-2}$. %for CARS196 and SOP. 
%Smaller base learning rate $1e^{-3}$ is better for CUB as it overlaps a little with ImageNet and contains fewer images~\cite{krause20133d,wang2017deep}. 
%On CARS and CUB, the training procedure converges at 10$k$ iterations, while 16$k$ iterations for SOP. 
%We set hyper-parameters empirically as follows $m=0.4, T_n=10 \text{ and } \alpha=1.2$?    
In each mini-batch, we randomly sample $C$ classes and $K$ images per class. 
We use a single Tesla V100 to train the large dataset SOP, while a single GTX 1080 Ti to train other datasets. 
To leverage the computational resources, we set $C=60, K =3$ on SOP while $C=22, K=3$ on the other three datasets. Thus $N_c=K=3, \forall c$, the batch size $N=180$ on SOP while $N=66$ on others. 
%On SOP, there are 2 positive images and 177 negative images in the ranked list corresponding to each query, which simulates the global set-based similarity structure. More precisely, only a few matching examples exist in a large gallery.
%In most cases, there are only a few positives, 
%Therefore 

We do not manually optimise the hyper-parameters extensively, i.e., we use `RLL-Simpler' to compare with existing methods, so that we only need to optimise $m$ and $T_n$.  
In addition, we simply fix $T_n$ throughout the training process. 
We only explore the dynamic weighting scheme when studying the critical learning periods in DML. 

In the ablation study, we use the full version of RLL for more comprehensive study. 
Moreover, in those experiments, we use the simpler data augmentation for faster convergence since our focus is analysing the key components, instead of comparing with existing methods. 
Specifically, we crop a random size of $224\times 224$ from the resized image ($256\times 256$), and horizontally flip the crop with a probability of 0.5. 

%we always set $T_p=0$ by default when there is no explicit declaration. 

%
%\noindent
%\subsubsection{Reproducibility}
For the ease of reproducing our results, 
%Our method is implemented  in the Caffe deep learning framework \cite{jia2014caffe}.
Our source code and the training scripts of all datasets are publicly available online: %https://github.com/XinshaoAmosWang/Ranked-List-Loss-for-DML. 
\url{https://github.com/XinshaoAmosWang/Ranked-List-Loss-for-DML}.
%We will release the source code and trained models for the ease of reproduction. 

%%%%%%%%%%%%%%%%%%%%%%%%%%%%%%%%
\begin{table}[!t]
	\caption{
		Comparison with the state-the-of-art methods on SOP. 
		XBM \cite{wang2020cross} is marked with `*' because it exploits information across mini-batch tasks.
		The `--' denotes the corresponding results are not reported in the original paper.
		The embedding dimension of different approaches is shown for the sake of clarity.
		%a more proper comparison. 
		%\\
		%For a fair comparison, all reported results have an embedding size of 64, which is the common setting and more challenging than a larger embedding size.  
		%\\
		%For a fair comparison, RLL-H denotes single-level embedding, i.e., the high-level embedding. RLL-(L,M,H) denotes multilevel embedding by concatenating the low-level, mid-level and high-level embeddings.
		%Only the R@1 results of RLL-H are bolded. 
		%and used for a fair comparison, because its architecture and embedding size are exactly the same as other baselines.
	}
	\label{table:SOTA_SOP}
	\centering
	%\vspace{-2pt}
	\setlength{\tabcolsep}{6pt} % Default value: 6pt
	\vspace{-0.1cm}
	\begin{tabular}{lcccc}
		\toprule
		&  
		& \multicolumn{3}{c}{SOP} \\ 
		\cmidrule(r){3-5}
		& Dim. & R@1 & R@10 & R@100 \\
		\midrule
		
		Lifted Struct \cite{song2016deep} &64 & 62.5 & 80.8 & 91.9 \\

		N-pair-mc \cite{sohn2016improved} & 64 &66.4 & 83.2 & 93.0 \\
		
		Triplet Semihard \cite{schroff2015facenet} &  64 & 66.7 & 82.4 & 91.9 \\
		
		Struct Clust \cite{song2017deep} & 64 &67.0 & 83.7 & 93.2   \\
		
		Spectral Clust \cite{law2017deep} & 512 &67.6 & 83.7 & 93.3 \\
		
		Proxy NCA \cite{movshovitz2017no} &64 & 73.7 & -- & --  \\
		HTL \cite{ge2018deep} & 512 &{74.8} & {88.3} & {94.8} \\
		
		Circle Loss \cite{sun2020circle} & 512 &{78.3} & {90.5} & {96.1}\\
		
		% When net is GoogLeNet V2
		XBM* \cite{wang2020cross} &512 & 79.5& 90.8& 96.1  \\
		
		\hdashline\\[-0.25cm]
		
		% ResNet-50
		%Divide \& Conquer \cite{sanakoyeu2019divide} & {75.9} & {88.4} & {94.9} \\
		
		Multi-Simi \cite{wang2019multi} & 64 & 74.1 & 87.8 & 94.7 \\
		Multi-Simi \cite{wang2019multi} & 128 & 76.6 & 89.2 & 95.2 \\
		
		Multi-Simi \cite{wang2019multi} & 512 &{78.2} & {90.5} & {96.0} \\
		
		\hdashline\\[-0.25cm]
		
		SoftMax$_{norm}$ \cite{qian2019softtriple} &64 & 75.9 &  88.8 &  95.2\\
		SoftMax$_{norm}$ \cite{qian2019softtriple} &512 & 78.0 &90.2 &96.0\\
		
		\hdashline\\[-0.25cm]
		
		SoftTriple \cite{qian2019softtriple} & 64 & 76.3 & 89.1 & 95.3\\
		
		SoftTriple \cite{qian2019softtriple} & 512 &78.3 &90.3 &95.9\\
		
		%\hdashline\\[-0.25cm]
		%\hdashline\\[-0.25cm]

		\midrule
		%\midrule
		%RLL-Simpler-L  & 64 & 74.9 & 87.9 & 94.5  \\
		
		%RLL-Simpler-M  & 64 & 75.7 & 88.7 & 95.1 \\
		
		RLL-Simpler-H  & 64 & {74.8} & {88.4} & {95.1} \\
		
		RLL-Simpler-(L,M,H) & $192=64\times$3 & {78.6} & {90.4}& {95.9} \\
		%\midrule
		
		\hdashline\\[-0.25cm]
		%RLL-Simpler-L  & 128 & 75.6 & 88.5 & 94.7 \\
		%RLL-Simpler-M  & 128 & 76.1 & 89.5 & 95.4 \\
		RLL-Simpler-H  & 128 & 75.0 & 88.7& 95.2 \\
		RLL-Simpler-(L,M,H) & $384=128\times$3 & 79.3 & 91.3 & 96.3\\
		%\midrule
		\hdashline\\[-0.25cm]
		%RLL-Simpler-L  & 512 & 76.4 & 89.0 & 95.0 \\
		%RLL-Simpler-M  & 512 & 76.9 & 89.7 & 95.5 \\
		RLL-Simpler-H  & 512 & 75.9 & 89.1 & 95.4 \\
		RLL-Simpler-(L,M,H) & $1536=512\times$3 & {80.0}  & {91.4} & {96.4} \\
		\bottomrule
	\end{tabular}

%	\begin{tablenotes}
%		%\item[a] $^\#$:    
%		\item[b] $^*$:  
%		Although the results of SoftMax$_{norm}$  and SoftTriple are better, they are theoretically non-scalable to extremely large datasets because they use multiple proxies to represent one class.
%		%because they exploit multiple proxies to represent one class.  \\
%		
%		%\item[a] {$^\#$: XBM exploits extra information across mini-batch tasks.}
%		
%		%\item[c]  In the bottom block,  
%	\end{tablenotes}
\end{table}
\begin{table}[!t]
	\caption{
		Comparison with the state-the-of-art methods on In-shop Clothes dataset. 
		XBM \cite{wang2020cross} is marked with `*' because it exploits information across mini-batch tasks.
		%XBM exploits extra information across mini-batch tasks.
		%We display the embedding dimension of different approaches for the sake of clarity.
		%The evaluation settings follow Table~\ref{table:SOTA_SOP}. 
		%The `--' denotes the corresponding results are not reported in the original paper. 
		%For a fair comparison, we compare with the reported results with an embedding size of 128, which is the common setting on In-shop Clothes. 
		%We use RLL-H to fairly compare with other baselines.    
		%Only the best R@1 result is bolded. 
	}
	\label{table:SOTA_InshopClothes}
	\centering
	%\vspace{-2pt}
	\setlength{\tabcolsep}{2.0pt} % Default value: 6pt
	\vspace{-0.1cm}
	\begin{tabular}{lccccccc}
		\toprule
		&& \multicolumn{6}{c}{In-shop Clothes} \\ 
		\cmidrule(r){3-8}
		& Dim. &  R@1 & R@10 & R@20 & R@30 & R@40 & R@50\\
		\midrule
		FashionNet  \cite{liu2016deepfashion} & 4096  &  53.0 & 73.0 & 76.0 & 77.0 & 79.0 & 80.0 \\
		
		HTL \cite{ge2018deep} & 128 & {80.9} & {94.3} & {95.8} & 97.2 & 97.4 & 97.8\\
        \hdashline\\[-0.25cm]
		Multi-Simi \cite{wang2019multi} & 128 & {88.0} & {97.2} & {98.1} & 98.5 & 98.7 & 98.8 \\
		Multi-Simi \cite{wang2019multi} & 512 & {89.7} & {97.9} & {98.5} & 98.8 & 99.1 & 99.2\\
        \hdashline\\[-0.25cm]
		
		% When net is GoogLeNet V2
		XBM*\cite{wang2020cross} & 512 & 89.9&  97.6& 98.4 &98.6 &98.8 &98.9 \\
		
		%\midrule
		\midrule
		
		%RLL-Simpler-L  & 128 & 85.2&96.3&97.5&98.0&98.3&98.5 \\
		%RLL-Simpler-M  & 128 & {88.7}&97.2&98.1&98.4&98.7&98.9 \\
		RLL-Simpler-H  & 128 & 88.5 & 97.1 & 98.0&98.4&98.6&98.7 \\
		
		RLL-Simpler-(L,M,H) & 128$\times$3 & {89.9}&97.6&98.3&98.7&98.9&99.0\\

		\bottomrule
	\end{tabular}

\end{table}
\begin{table}[!t]
	\caption{Comparison with the state-of-the-art methods on CARS196 in terms of Recall@$K$ (\%). 
		%We display the embedding dimension of different approaches for the sake of clarity.
		For an exact comparison, all the
		reported results have an embedding size of 64.
		% except for the concatenation.
		%The comparison settings follow Table~\ref{table:SOTA_SOP}.
		%We only report those results with an embedding size of 64 for a fair comparison.  
		%For a comparison, all the reported results have an embedding size of 64.  
		%The results of RLL-H are bolded. 
	}
	\label{table:SOTA_CARS}
	%\vspace{-2pt}
	\centering
	\setlength{\tabcolsep}{7.2pt} % Default value: 6pt
	\vspace{-0.1cm}
	\begin{tabular}{lccccc}
		\toprule
		&& \multicolumn{4}{c}{CARS196}  \\ 
		\cmidrule(r){3-6}
		&Dim.& R@1 & R@2 & R@4 & R@8  \\
		\midrule
		Triplet Semihard \cite{schroff2015facenet} & 64 & 51.5 & 63.8 & 73.5 & 82.4   \\
		Lifted Struct \cite{song2016deep} & 64 & 53.0 & 65.7 & 76.0 & 84.3   \\
		
		$N$-pair-mc \cite{sohn2016improved} & 64& 53.9 & 66.8 & 77.8 & 86.4  \\

		Struct Clust \cite{song2017deep}& 64 & 58.1 & 70.6 & 80.3 & 87.8 \\
		
		%Spectral Clust \cite{law2017deep} & 196 & 73.1 & 82.2 & 89.0 & 93.0  \\

		Proxy NCA \cite{movshovitz2017no}& 64 & 73.2 & 82.4 & 86.4 & 88.7 \\
		%HTL \cite{ge2018deep} & 512 & {81.4} & {88.0} & {92.7} & {95.7} \\ 
		%Circle Loss \cite{sun2020circle} & 512 & 83.4 & 89.8  & 94.1 & 96.5 \\
		
		%\hdashline\\[-0.25cm]
		%\midrule
		
		Multi-Simi \cite{wang2019multi} & 64 & 77.3 & 85.3 & 90.5 & 94.2  \\
		%Multi-Simi \cite{wang2019multi} & 512 & 84.1 & 90.4 & 94.0 & 96.5 \\
		
		%\midrule
		\hdashline\\[-0.25cm]
		
		SoftMax$_{norm}$ \cite{qian2019softtriple} & 64 & 76.8 &85.6 &91.3 &95.2  \\
		
		%SoftMax$_{norm}$ \cite{qian2019softtriple} & 512 & 83.2 &89.5 &94.0 &96.6 \\
		
		%\midrule
		%\hdashline\\[-0.25cm]
		
		SoftTriple \cite{qian2019softtriple} & 64 & 78.6 &86.6 &91.8 &95.4  \\
		
		%SoftTriple \cite{qian2019softtriple} & 512& 84.5 &90.7 &94.5 &96.9 \\
		\hdashline\\[-0.25cm]
		%\midrule

		{RLL-Simpler-H} & 64 & 73.7 & 82.4 & 88.8 & 93.3 \\
		
		%{RLL-Simpler-(L,M,H)} & 64$\times$3 & {81.3} & {88.5} & {93.4} & {96.3} \\
		%0.813, rank2: 0.885, rank4: 0.934, rank8: 0.963

		%\hdashline\\[-0.25cm]
		
		%{RLL-Simpler-H} & 512 &   {74.0} & {83.6} & {90.1} & {94.1}  \\

		%{RLL-Simpler-(L,M,H)} & 512$\times$3 & {82.1} & {89.3} & {93.7} & {96.7} \\
		
		\bottomrule
	\end{tabular}
	
	%\vspace{-0.3cm}
\end{table}
%%%%%%%%%%%%%%%%%%%%%%%%%%%%%%%%%

%%%%%%%%%%%%%%%%%%%%%%%%%%%%%%%%
\begin{table}[!t]
	\caption{Comparison with the state-of-the-art methods on CUB-200-2011 in terms of Recall@$K$ (\%). 
		All the displayed results
		have an embedding size of 64 for an exact comparison.
		%The comparison settings follow Table~\ref{table:SOTA_SOP}. 
		%`\#Classes' denotes the number of classes. 
		%The embedding dimension of different approaches is shown.   
		%RLL-H's results are bolded. 
	}
	\label{table:SOTA_CUB}
	%\vspace{-2pt}
	\centering
	\setlength{\tabcolsep}{7.2pt} % Default value: 6pt
	\vspace{-0.1cm}
	\begin{tabular}{lccccc}
		\toprule
		&& \multicolumn{4}{c}{CUB-200-2011} \\ 
		\cmidrule(r){3-6}
		&Dim.& R@1 & R@2 & R@4 & R@8 \\
		\midrule
		Triplet Semihard \cite{schroff2015facenet} & 64 & 42.6 & 55.0 & 66.4 & 77.2 \\
		Lifted Struct \cite{song2016deep} & 64 & 43.6 & 56.6 & 68.6 & 79.6 \\
		
		$N$-pair-mc \cite{sohn2016improved} & 64& 45.4 & 58.4 & 69.5 & 79.5  \\
		
		Struct Clust \cite{song2017deep}& 64 & 48.2 & 61.4 & 71.8 & 81.9  \\
		
		%Spectral Clust \cite{law2017deep} & 200 & 53.2 & 66.1 & 76.7 & 85.3  \\

		Proxy NCA \cite{movshovitz2017no}& 64 & 49.2 & 61.9 & 67.9 & 72.4 \\
		%HTL \cite{ge2018deep} & 512 & 57.1 & 68.8 & 78.7 & 86.5  \\ 
		%Circle Loss \cite{sun2020circle} & 512 & 66.7 & 77.4 & 86.2 & 91.2  \\
		%\hdashline\\[-0.25cm]
		Multi-Simi \cite{wang2019multi} & 64 & 57.4 & 69.8 & 80.0 & 87.8 \\
		%Multi-Simi \cite{wang2019multi} & 512 & 65.7 & 77.0 &86.3 &91.2  \\
		
		\hdashline\\[-0.25cm]
		SoftMax$_{norm}$ \cite{qian2019softtriple} & 64 & 57.8 &70.0 &80.1 &87.9 \\

		%SoftMax$_{norm}$ \cite{qian2019softtriple} & 512 & 64.2 &75.6& 84.3& 90.2\\
		
		SoftTriple \cite{qian2019softtriple} & 64 & 60.1 &71.9 &81.2 &88.5 \\
		
		%SoftTriple \cite{qian2019softtriple} & 512& 65.4 &76.4 &84.5 &90.4 \\
		
		%\midrule
		
		\hdashline\\[-0.25cm]
		{RLL-Simpler-H} & 64 &  	
		56.7 & 68.1 & 77.7 & 85.8\\
		
		%{RLL-Simpler-H} & 512 &   		{57.4} & {69.7} & {79.2} & {86.9} \\

		%{RLL-Simpler-(L,M,H)} & 512$\times$3 & {63.0} & {74.2} & {82.8} & {89.2} \\
		%0.630, rank2: 0.742, rank4: 0.828, rank8: 0.892
		\bottomrule
	\end{tabular}
	
	%\vspace{-0.3cm}
\end{table}
%%%%%%%%%%%%%%%%%%%%%%%%%%%%%%%%%

\begin{table}[!t]
	\caption{
		Comparison with the state-the-of-art methods on SOP when the deep net architecture is ResNet-50 \cite{he2016deep}.         
		For an exact comparison, the embedding size is 128 for all reported results.        
		Since ResNet-50 has a high-level embedding only, we omit `H' for the sake of brevity. 
	}    
	\label{table:SOTA_SOP_RESNET}
	\centering
	%\vspace{-2pt}
	\setlength{\tabcolsep}{9.3pt} % Default value: 6pt
	\vspace{-0.1cm}
	\begin{tabular}{lcccc}
		\toprule
		&  
		& \multicolumn{3}{c}{SOP} \\ 
		\cmidrule(r){3-5}
		& Dim. & R@1 & R@10 & R@100 \\
		\midrule
		
		Margin \cite{wu2017sampling} & 128 & 72.7 & 86.2 & 93.8  \\

		Divide \& Conquer \cite{sanakoyeu2019divide} & 128 &  75.9 &  88.4 & 94.9\\
		
		FastAP \cite{cakir2019deep} &  128 & 73.8 &  88.0 &  94.9 \\
		
		MIC \cite{roth2019mic} & 128 & 77.2 & 89.4 & 95.6   \\

		%\midrule
		\hdashline\\[-0.25cm]
		
		RLL-Simpler  & 128 & \textbf{{78.7}} & \textbf{{91.1}} & \textbf{{96.4}} \\
		\bottomrule
	\end{tabular}  
\end{table}

\subsection{Comparison with Recent Baselines}
\label{section:state_of_the_art}
%%%%%%%%%%%%%%%%%%%%%%%%%%%%%%%%%%%%%%%%%%%%%%%%%%%%%%%
%The NMI score measures the quality of clustering while Recall@$K$ denotes the performance of image retrieval. Both metrics are commonly used to evaluate the ability of the learned embedding function for extracting discriminative features.  
%

%Except for Spectral Clust and Proxy NCA, the results of all baselines (Triplet Semihard, Lifted Struct, $N$-pair-mc, Struct Clust) come from Struct Clust where all methods are implemented and tested under the same setting. Spectral Clust, Proxy NCA and our method follow the same settings as in Struct Clust

\noindent
\textbf{Competitors.}
We compared our method with the following methods which are implemented and tested under the same settings: Triplet Semihard, Lifted Struct, N-pair-mc, Struct Clust, Spectral Clust, and Proxy NCA. 
%\footnote{
%	Some methods \cite{harwood2017smart,wang2017deep,ustinova2016learning,duan2018deep,lin2018deep,suh2019stochastic} use GoogLeNet V1 \cite{szegedy2015going}. Some others, e.g., the margin loss \cite{wu2017sampling}, FastAP \cite{cakir2019deep}, MIC \cite{roth2019mic} and Divide \& Conquer \cite{sanakoyeu2019divide}, apply ResNet50 \cite{he2016deep}. For a fair comparison, they are not compared in the table. Additionally, we {do not compare with ensemble models}~\cite{yuan2017hard,opitz2017bier,kim2018attention,xuan2018deep}.
%}.
%
These methods have been described in Section~\ref{section:preliminary} except for Triplet Semihard \cite{schroff2015facenet} which mines semihard negative examples to improve the conventional triplet loss and is reimplemented in Struct Clust\cite{song2017deep} with GoogLeNet V2. 
HTL \cite{ge2018deep} exploits the hierarchical similarity structure among different classes and merges similar classes recursively, i.e., building the global class-level hierarchical tree by using all original classes as leaves and updates the tree after every epoch.  HTL \cite{ge2018deep} is computationally expensive and unscalable to extremely large datasets. 
Additionally, SoftMax$_{norm}$  and SoftTriple \cite{qian2019softtriple} are theoretically non-scalable to extremely large dataset because they use multiple proxies to represent one class. 
%In addition, 
XBM \cite{wang2020cross} exploits extra information across mini-batch tasks.
%, so that it is not benchmarked.
% 
%Additionally, 
%We {do not compare with ensemble models}~\cite{yuan2017hard,opitz2017bier,kim2018attention,xuan2018deep}.
%We remark that 
%Some methods \cite{harwood2017smart,wang2017deep,ustinova2016learning,duan2018deep,lin2018deep,suh2019stochastic} use GoogLeNet V1 \cite{szegedy2015going}, while some others, e.g., the margin loss \cite{wu2017sampling}, FastAP \cite{cakir2019deep}, MIC \cite{roth2019mic} and Divide \& Conquer \cite{sanakoyeu2019divide}, apply ResNet50 \cite{he2016deep}. 
%
Some other methods, e.g., Margin \cite{wu2017sampling}, Divide \& Conquer \cite{sanakoyeu2019divide}, FastAP \cite{cakir2019deep} and MIC \cite{roth2019mic}, use ResNet-50 \cite{he2016deep} as the backbone network. 
To compare with them fairly, we also apply RLL-Simper to train ResNet-50.  
%Since ResNet-50 has a high-level embedding only, we omit `H' for the sake of brevity 
%when displaying the results of ResNet-50 in Table~\ref{table:SOTA_SOP_RESNET}.
%As a result, they are not compared in the table. 
%We choose not to train GoogLeNet V2 using those methods because of two reasons: (1) We need to tune their hyperparameters to obtain good performance when training GoogLeNet V2; (2) 
%But t
%Although they are not compared in the table, the majority of them have been compared and outperformed in the most recent algorithms: Multi-Simi \cite{wang2019multi}, SoftMax$_{norm}$ and SoftTriple, Circle Loss \cite{sun2020circle}, and  XBM.

%Although HTL \cite{ge2018deep} achieves good performance on small datasets but is worse than ours on the large dataset SOP.
% improve the traditional triplet loss. 

%Although HTL~\cite{ge2018deep} also uses GoogLeNet V2, we do not benchmark it because it builds the global class-level hierarchical tree by using all original classes as leaves and updates the tree after every epoch, thus being  very computationally expensive and unscalable.
%%%%%%%%%%%%%%%%%%%%%%%%%%%%%%%%%%%%%%%%%%

%%%%%%%%%%%%%%%%%%%%%%%%%%%%%%%%%%%%%%%%%%%%%%
%
\noindent
\textbf{Result analysis.} The comparisons between our method and existing competitors on four datasets are presented in Tables~\ref{table:SOTA_SOP}, \ref{table:SOTA_InshopClothes}, \ref{table:SOTA_CARS}, \ref{table:SOTA_CUB}, and \ref{table:SOTA_SOP_RESNET}. 
%
%
%As stated in Section~\ref{section:implemtation_details}, we have three fully connected layers in GoogLeNet V2. We report two sets of results. For a fair comparison, we report the results of the high-level embedding, which is denoted as RLL-H. In addition, 
%We empirically find that the multilevel embedding (RLL-(L,M,H) in short) by concatenating the low-level, mid-level and high-level embeddings can achieve better performance.
%For a fair comparison, on every dataset, we report the results of RLL-Simpler-H whose network architecture and embedding size are exactly the same as other reported baselines.  
%
All reported approaches use the high-level embedding when `L', `M', `H' or `(L, M, H)' is not explicitly marked.
From these tables, we have the following observations:   
\begin{itemize}
	\item On the two larger datasets, RLL-Simpler has the state-of-the-art performance, as shown in Tables \ref{table:SOTA_SOP}, \ref{table:SOTA_InshopClothes} and \ref{table:SOTA_SOP_RESNET}. 
	
	\item On the two smaller datasets, except for the theoretically non-scalable SoftMax$_{norm}$ and SoftTriple, only Multi-Simi is better than ours. 
	However, technically, the weighting scheme of  Multi-Simi considers multiple metrics, thus being more complex than ours. 
	%\item There are some other interesting results on two larger datasets: 
	%(1) RLL-L and RLL-M use shallower subnetworks, while perform similarly or even better than RLL-H;
	%RLL-M is better than RLL-H although its network depth is slightly shallower; RLL-L's result is also promising although its depth is much shallower. 
	%(2) We simply increase the embedding size by feature concatenation, and outperform other reported results of an even higher embedding size. 
\end{itemize}
%While performs a bit worse on the smaller ones CARS196 and CUB-200-2011. 

We remark: (1) in the general settings of metric learning, the training classes are disjoint with testing classes, which means the training set can be infinitely large. Therefore, larger datasets, e.g, SOP and In-shop Clothes, are better to test on;  
(2) CARS196 and CUB-200-2011 are significantly sensitive to the minor changes of training details as mentioned in \cite{wang2019multi,qian2019softtriple,musgrave2020metric}.

\subsection{On the network depth of an embedding function}
%\label{section:ablation}

In this subsection, we study the network depth of an embedding function and the concatenation of embedding functions. The results on SOP and In-shop Clothes are displayed in Tables~\ref{table:SOP_network_depth} and \ref{table:Inshop_network_depth}, respectively.  
RLL-Simpler-(L,M,H) denotes the multilevel embedding by concatenating the low-level, mid-level and high-level embeddings. Others are analogous. 
We study and report them because from the practical perspective, RLL-Simpler-L and RLL-Simpler-M are of smaller network depth, while perform similarly or even better than RLL-Simpler-H. 
Firstly, this indicates that very deep networks are not necessarily better in practice. 
This information is of great value in applications where smaller networks with faster computational speed are preferred.
%Those additional results are valuable for many real-world applications.   
Secondly, this motivates us to exploit multi-layer embeddings well.
All of them are discriminative, and exploit non-identical knowledge to measure the distance between data points since they are of different depth. We obtain promising performance by simply concatenating them.  Therefore, there is an open future research lead to better exploit multi-layer embeddings in deep metric learning, without the need to train multiple learners~\cite{yuan2017hard,opitz2017bier,kim2018attention,xuan2018deep}.

%%%%%%%%%%%%%%%%%%%%%%%%%%%%%%%%%
\begin{table}[!t]
	\caption{
		Exploration on the network depth of an 
		embedding function and the concatenation of embedding functions on SOP. 
		%Additionally, the bottom block shows the recently reported results using a higher embedding size, except for the results in Table~\ref{table:SOTA_SOP}.
	}
	\label{table:SOP_network_depth}
	\centering
	%\vspace{-2pt}
	\setlength{\tabcolsep}{7pt} % Default value: 6pt
	\vspace{-0.1cm}
	\begin{tabular}{lcccc}
		\toprule
		&  
		& \multicolumn{3}{c}{SOP} \\ 
		\cmidrule(r){3-5}
		& Dim. & R@1 & R@10 & R@100 \\
		\midrule
		
		RLL-Simpler-L  & 64 & 74.9 & 87.9 & 94.5  \\
		
		RLL-Simpler-M  & 64 & 75.7 & 88.7 & 95.1 \\
		
		RLL-Simpler-H  & 64 & {74.8} & {88.4} & {95.1} \\
		
		RLL-Simpler-(L,M) & $128=64\times$2 & 78.3&90.2&95.8 \\
		
		RLL-Simpler-(L,H) & $128=64\times$2 & 78.4&90.1&95.8 \\
		
		RLL-Simpler-(M,H) & $128=64\times$2 & 76.9 & 89.5&95.5 \\
		
		RLL-Simpler-(L,M,H) & $192=64\times$3 & 78.6 & 90.4&95.9 \\

		%XBM \cite{wang2020cross} $^\#$ &512 & 79.5& 90.8& 96.1 \\
		
		\bottomrule
	\end{tabular}
	%\begin{tablenotes}
		%\item[a] $^\#$:    
		%\item[b] $^*$:  SoftMax$_{norm}$  and SoftTriple are theoretically non-scalable to extremely large datasets.
		%because they exploit multiple proxies to represent one class.  \\
		
		%\item[a] {$^\#$: XBM exploits extra information across mini-batch tasks.}
		
		%\item[c]  In the bottom block,  
	%\end{tablenotes}
\end{table}
%%%%%%%%%%%%%%%%%%%%%%%%%%%%%%%%%%

%%%%%%%%%%%%%%%%%%%%%%%%%%%%%%%%
\begin{table}[!t]
	\caption{
		Exploration on the network depth of an 
		embedding function and the concatenation of embedding functions on In-shop Clothes dataset. 
	%	In addition, the recently reported results using a higher embedding size are displayed in the bottom block, except for the results in Table~\ref{table:SOTA_InshopClothes}.
	}
	\label{table:Inshop_network_depth}
	\centering
	%\vspace{-2pt}
	\setlength{\tabcolsep}{2.0pt} % Default value: 6pt
	\vspace{-0.1cm}
	\begin{tabular}{lccccccc}
		\toprule
		&& \multicolumn{6}{c}{In-shop Clothes} \\ 
		\cmidrule(r){3-8}
		&Dim.&  R@1 & R@10 & R@20 & R@30 & R@40 & R@50\\
		\midrule
		%FashionNet  \cite{liu2016deepfashion} & 4096  &  53.0 & 73.0 & 76.0 & 77.0 & 79.0 & 80.0 \\

		%\midrule
		
		%Multi-Simi \cite{wang2019multi} & 512 & {89.7} & {97.9} & {98.5} & 98.8 & 99.1 & 99.2\\
		
		% When net is GoogLeNet V2
		%XBM \cite{wang2020cross} $^\#$ \tnote{b} & 512 & {89.9}&  97.6& 98.4 &98.6 &98.8 &98.9 \\
		%\midrule
		%\midrule
		
		RLL-Simpler-L  & 128 & 85.2&96.3&97.5&98.0&98.3&98.5 \\
		
		RLL-Simpler-M  & 128 & {88.7}&97.2&98.1&98.4&98.7&98.9 \\
		
		RLL-Simpler-H  & 128 & {88.5} & 97.1 & 98.0&98.4&98.6&98.7 \\

		RLL-Simpler-(L,M) & 128$\times$2 & 89.3 & 97.5 & 98.2 & 98.6 & 98.8 & 99.0 \\
		
		RLL-Simpler-(L,H) & 128$\times$2 & 89.4 & 97.5 & 98.3 & 98.6 & 98.8 & 98.9 \\

		RLL-Simpler-(M,H) & 128$\times$2 & 89.4&97.4&98.2&98.5&98.7&98.9\\
		
		%$^\&$
		RLL-Simpler-(L,M,H) & 128$\times$3 & {89.9}&97.6&98.3&98.7&98.9&99.0\\
		%\midrule
		
		%FashionNet  \cite{liu2016deepfashion} & 4096  &  53.0 & 73.0 & 76.0 & 77.0 & 79.0 & 80.0 \\

		%Multi-Simi \cite{wang2019multi} & 512 & {89.7} & {97.9} & {98.5} & 98.8 & 99.1 & 99.2\\
		
		% When net is GoogLeNet V2
		%XBM \cite{wang2020cross} $^\#$ \tnote{b} & 512 & {89.9}&  97.6& 98.4 &98.6 &98.8 &98.9 \\
		%\midrule
		
		%RLL-L  & 512 & 85.9&96.7&97.8&98.3&98.6&98.7 \\
		%RLL-M  & 512 & 87.6&96.9&97.9&98.3&98.5&98.7 \\
		%RLL-H  & 512 & 87.1&96.7&97.8&98.3&98.5&98.7 \\
		%RLL-(L,M,H) & 512x3 & 89.6& 97.5&98.3&98.6&98.9&99.0 \\
		\bottomrule
	\end{tabular}
%	\begin{tablenotes}
%		%\item[b] {$^\#$: We use RLL-H to fairly compare with other baselines.}
%		\item[a] {$^\#$: XBM exploits  extra information across mini-batch tasks.}
%		%\item $^\&$: We increase the embedding size by simply concatenating multi-layer representations, instead of training another model with a larger embedding dimension.   The results are shown in the bottom block. 
%	\end{tablenotes}
\end{table}
%%%%%%%%%%%%%%%%%%%%%%%%%%%%%%%%%

\subsection{Mining Non-trivial Pairs}
%\label{section:ablation}

As presented in Section \ref{section:RLL}, for each query, RLL  mines examples which violate the pairwise constraint with respect to the query. Specifically, we mine negative examples whose distance is smaller than $\alpha$ in Eq.~(\ref{equation:RLL_N}). Simultaneously, we mine positive examples whose distance is larger than $\alpha-m$ in Eq.~(\ref{equation:RLL_P}). 
As a result, a margin $m$ is established between negative and positive examples in each ranked list. 
%
%Since the sample mining range, the constraint parameters $\alpha, m$,  
What examples are mined 
is determined by $\alpha$ and $m$. 
We conduct experiments on the large dataset SOP to analyse the influence of $\alpha$. 
Regarding {$m$}, we study it in section~\ref{subsection:hyperball_regularisation}.

%%%%%%%%%%%%%%%%%%%%%%%%%%%%%%%%
\begin{table}[!t]
	\caption{The impact of $\alpha$ on the SOP. In all experiments, $m=0.4, T_n=10$.  
	}
	\label{table:negative_constraint}
	\centering
	%\vspace{-2pt}
	\vspace{-0.1cm}
	\begin{tabular}{ccccc}
		\toprule
		$m=0.4, T_n=10$ & R@1 & R@10 & R@100 \\
		\midrule
		$\alpha=1.4$ & 76.2 & 89.4 & 95.6 
		\\
		$\alpha=1.2$ & {79.8} & {91.3} & {96.3} \\
		$\alpha=1.0$ & 78.7 & 90.5 & 95.9 
		\\
		\bottomrule
	\end{tabular}
\end{table}
%%%%%%%%%%%%%%%%%%%%%%%%%%%%%%%%%\\
%%%%%%%%%%%%%%%%%%%%%%%%%%%%%%%%
%\begin{table}[!b]
%	\caption{The impact of the distance margin $m$ between negative and positive examples. The Recall@$K$ (\%) results on SOP  are shown with $\alpha=1.2, T_n=10$ in all experiments. 
%	}
%	\label{table:margin}
%	\centering
%	\vspace{-2pt}
%	\begin{tabular}{lcccc}
%		\hline
%		$\alpha=1.2, T=10$  & R@1 & R@10 & R@100 \\
%		\hline
%		$m=0$ & 76.1 & 89.8 & 95.7 \\
%		$m=0.2$ & 79.0 & 91.2 & \textbf{96.3} \\
%		%$m=0.2,5k$ & 79.4 & 91.4 & 96.4 \\
%		$m=0.4$ & \textbf{79.8} & \textbf{91.3} & \textbf{96.3} \\
%		$m=0.6$ & 79.2 & 90.6 & 96.0 \\
%		%$m=0.8$ & 79.0 & 90.5 & 95.9 \\
%		%$m=1.0$ & 79.1 & 90.5 & 95.9 \\
%		$m=1.2$ & 79.1 & 90.5 & 95.8 \\
%		\hline
%	\end{tabular}
%\end{table}
%%%%%%%%%%%%%%%%%%%%%%%%%%%%%%%%%\\
\noindent
\textbf{Impact of $\alpha$}. To study the impact of $\alpha$, we set the temperature $T_n=10$ and the margin $m=0.4$ in all experiments. The results are presented in Table~\ref{table:negative_constraint}. We observe that a proper negative constraint $\alpha$ is important for RLL to learn discriminative embeddings. This is consistent with our intuition as $\alpha$ controls how much the negative examples are pushed away.    

\subsection{Weighting Negative Pairs}
%There are still a lot of negative examples after mining. 
In this section, we conduct experiments to evaluate the influence of $T_n$ for weighting negative examples in Eq.~(\ref{equation:w_ij}). We fix $m=0.4$ and $\alpha=1.2$ in all experiments. The temperature parameter $T_n \ (T_n>0)$  controls the slope of weighting. The results are presented in Table~\ref{table:T}. 
We observe that:
\begin{itemize}
	\item When $T_n=0$, RLL treats all non-trivial negative examples equally, i.e., no weighting is applied. The Recall@$1$ result is 78.8\%, which is only 1\% lower than the best one using proper weighting. This demonstrates the  superiority of RLL even without weighting.
	
	%\vspace{-7pt}
	\item The performance changes slightly as $T_n$ does. The performance gap is around 1\% when $T_n$ ranges from 0 to 20.  In addition, the performance drops when $T_n$ is too large. This may be because `very' hard examples exist in the training data (e.g., outliers) \cite{schroff2015facenet,cui2016fine,wang2019deep}.
\end{itemize}

%%%%%%%%%%%%%%%%%%%%%%%%%%%%%%%%
\begin{table}[!t]
	\caption{Results of different $T_n$ on the SOP in terms of Recall@$K$ (\%). We fix $m=0.4, \alpha=1.2$ in all experiments.
	}
	\label{table:T}
	\centering
	%\vspace{-2pt}
	\vspace{-0.1cm}
	\begin{tabular}{ccccc}
		\toprule
		$m=0.4, \alpha=1.2$ & R@1 & R@10 & R@100 \\
		\midrule
		$T_n=0$ & 78.8 & 90.7 & 96.1  
		\\
		$T_n=5$ & 79.1 & 91.0 & 96.2  
		\\
		$T_n=10$ & {79.8} & {91.3} & {96.3} 
		\\
		$T_n=15$ & 79.3 & 90.9 & 96.0  
		\\
		$T_n=20$ & 78.6 & 90.5 & 95.7 
		\\
		%$T=30$ %& 74.5 & 87.9 & 94.3 
		%\\
		\bottomrule
	\end{tabular}
	\vspace{-0.25cm}
\end{table}
%%%%%%%%%%%%%%%%%%%%%%%%%%%%%%%%%\\
%\vspace{-5pt}

%\subsubsection{Independent Normalisation}

%%%%%%%%%%%%%%%%%%%%%%%%%%%%%%%%
%\begin{table}[!h]
%	\caption{The results of with/without independent normalisation (IN) on SOP in terms of Recall@$K$ (\%).
% and NMI (\%) 
%	}
%	\label{table:normalisation}
%	\centering	
%	\begin{tabular}{lcccc}
%		\hline
%		 & $K=1$ & $K=10$ & $K=100$ \\
%		\hline
%		with IN & \textbf{79.8} & \textbf{91.3} & \textbf{96.3} \\
%		without IN & 79.2 & 91.3 & 96.4 \\
%		\hline
%	\end{tabular}
%\end{table}
%%%%%%%%%%%%%%%%%%%%%%%%%%%%%%%%%\\

%%%%%%%%%%%%%%%%%%%%%%%%%%%%%%%%%%%%%%%%
\subsection{Weighting Positive Pairs}
In most cases where only a few positive data points exist, we only weight negative examples. 
%The reason is that 
For example, when $K=3$, given a query, there are only two positives for every query so that we treat them equally. 
We have tried differentiating them but the performance difference is neglectable. 
However, when multiple positives exist, it is quite natural to ask whether weighting positive examples helps. 
Therefore, following \cite{wang2019deep}, we empirically study weighting positive examples when there are many positive data points in a retrieved list. 

While fixing the batch size to be 180, we study two different cases: $K=6$ and $K=12$. Given a query, there are 5 and 11 positive instances out of 179 in its retrieved list, respectively. 
In each case, we choose 5 different $T_p$: 10, 5, 0, -5, -10. We remark that: 1) $T_p > 0$ denotes harder positives with a larger distance are emphasised while $T_p < 0$ represents easier positive pairs with a smaller distance are focused; 2) When the absolute value of $T_p$ is larger, e.g., 10 and -10, the relative weight between two instances is larger. As a result, the differentiation becomes more significant. 
The results are shown in Table~\ref{table:postives_weighting} and we discuss them as follows:     

\begin{itemize}
	\item When emphasising on easier positive examples ($T_p < 0$) or without weighting ($T_p=0$), the performance is similar. 
	\item When focusing on harder positive examples, the performance decreases as $T_p$ increases.   
	\item The results are generally consistent with \cite{wang2019deep,cui2016fine}. They find that emphasising on harder positive examples cannot preserve the intraclass similarity manifold structure, thus reducing the generalisation performance. 
\end{itemize}

%\subsubsection{Emphasising on Harder Positive Examples}
%
%When emphasising on harder positives, ...
%
%\subsubsection{Emphasising on Easier Positive Examples}
%
%When emphasising on easier positives, ...

%%%%%%%%%%%%%%%%%%%%%%%%%%%%%%%%
\begin{table}[!t]
	\caption{
		Results of weighting positive examples on the SOP. We fix other parameters, i.e., $m=0.4, \alpha=1.2 \text{ and } T_n = 10$.
	}
	\label{table:postives_weighting}
	\centering
	\setlength{\tabcolsep}{6pt} % Default value: 6pt
	%\vspace{-6pt}
	%\fontsize{8.8pt}{9.8pt}\selectfont
	\vspace{-0.1cm}
	\begin{tabular}{lcccc}
		\toprule
		\multirow{2}{*}{\makecell{Batch Size,\\Batch Content}} & 
		\multirow{2}{*}{\makecell{Weighting \\Positive Examples}} & %\multirow{2}{*}{\makecell{Emphasis\\Focus}} & \multirow{2}{*}{Model} & 
		\multicolumn{3}{c}{\makecell{Recall@$K$}} 
		%& \multicolumn{2}{c}{ \makecell{Accuracy  on\\ Training Sets (\%)} }
		\\
		\cmidrule{3-5}  
		&  & 1 & 10 & 100\\
		\cmidrule{1-5}
		\multirow{5}{*}{\makecell{$C=30, K =6,$\\$N=180=30\times6$}}
		&$T_p=$ 10&77.3 & 90.1 & 96.0 \\
		&$T_p=$ 5&78.5 & 90.7 & 96.2 \\
		&$T_p=$ 0&79.0 & 90.9 & 96.2 \\
		&$T_p=$ -5&78.8 & 90.7 & 95.9 \\
		&$T_p=$ -10&78.8 & 90.4 & 95.8 \\
		\cmidrule{1-5}
		\multirow{5}{*}{\makecell{$C=15, K =12,$\\$N=180=15\times12$}}
		&$T_p=$ 10&75.5 & 88.9 & 95.6 \\
		&$T_p=$ 5&76.5 & 89.4 & 95.8 \\
		&$T_p=$ 0&76.9 & 89.5 & 95.6 \\
		&$T_p=$ -5&76.9 & 89.2 & 95.3 \\
		&$T_p=$ -10&76.5 & 89.0 & 95.1 \\
		\bottomrule
	\end{tabular}
	%\vspace{-0.3cm}
\end{table}
%%%%%%%%%%%%%%%%%%%%%%%%%%%%%%%%%

%%%%%%%%%%%%%%%%%%%%%%%%%%%%%%%%
\begin{table}[!t]
	\caption{
		The impact of hypersphere diameter for preserving the intraclass variance on the SOP. We set $\alpha=1.2, T_n=10$ in all experiments.
	}
	\label{table:pn_margin}
	\centering
	\setlength{\tabcolsep}{6pt} % Default value: 6pt
	%\vspace{-6pt}
	%\fontsize{8.8pt}{9.8pt}\selectfont
	\vspace{-0.1cm}
	\begin{tabular}{lccccc}
		\toprule
		\multirow{2}{*}{\makecell{Batch Size,\\Batch Content}} & 
		\multirow{2}{*}{\makecell{Distance \\Margin $m$}} &
		\multirow{2}{*}{\makecell{Hypersphere \\Diameter\\$\alpha - m$}} & %\multirow{2}{*}{\makecell{Emphasis\\Focus}} & \multirow{2}{*}{Model} & 
		\multicolumn{3}{c}{\makecell{Recall@$K$}} 
		%& \multicolumn{2}{c}{ \makecell{Accuracy  on\\ Training Sets (\%)} }
		\\
		\cmidrule{4-6}  
		&  & & 1 & 10 & 100\\
		%
		%\cmidrule{1-6}
		\midrule
		\multirow{7}{*}{$180=60\times3$}
		&$m=0$ &1.2 & 76.1 & 89.8 & 95.7 \\
		&$m=0.2$&1.0 & 79.0 & 91.2 & {96.3} \\
		%$m=0.2,5k$ & 79.4 & 91.4 & 96.4 \\
		&$m=0.4$&0.8 & {79.8} & {91.3} & {96.3} \\
		&$m=0.6$&0.6 & 79.2 & 90.6 & 96.0 \\
		& $m=0.8$&0.4 & 79.0 & 90.5 & 95.9 \\
		& $m=1.0$&0.2 & 79.1 & 90.5 & 95.9 \\
		&$m=1.2$&0.0 & 79.1 & 90.5 & 95.8 \\
		
		\midrule
		\multirow{7}{*}{$180=30\times6$}
		&$m=0$ &1.2& 78.7 & 91.1 & {96.3} \\
		&$m=0.2$&1.0 & {79.1} & {91.2} & {{96.3}} \\
		&$m=0.4$&0.8 & {79.0} & {90.7} & {95.9} \\
		&$m=0.6$&0.6 & 78.3 & 89.7 & 95.4 \\
		&$m=0.8$&0.4 & 77.8 & 89.3 & 95.1 \\
		&$m=1.0$&0.2 & 76.2 & 88.0 & 94.3 \\
		&$m=1.2$&0.0 & 76.4 & 88.2 & 94.4 \\
		\bottomrule
	\end{tabular}
	%\vspace{-0.3cm}
\end{table}
%%%%%%%%%%%%%%%%%%%%%%%%%%%%%%%%%

\subsection{Hypersphere Regularisation}
\label{subsection:hyperball_regularisation}

The hypersphere diameter is an indicator of the intraclass variance. 
To study the impact of it, we fix $\alpha=1.2$ and $T_n=10$ while changing $m$. 
For more comprehensive exploration, we try two different settings:  $C=60, K=3$, and $C=30, K =6$.
We do experiments on the SOP dataset and present the results in Table~\ref{table:pn_margin}. 
We have two important observations: 
\vspace{-7pt}
\begin{itemize}
	%\item When $m>0$, RLL performs much better by around 3\% than $m=0$. It shows that the margin is important for improving the generalisation capability of RLL.
	
	%\vspace{-7pt}
	%\item The margin-based RLL is insensitive to the margin value. The performance difference is smaller than 1\% when $m$ ranges from 0.2 to 1.2.
	
	%\vspace{-7pt}
	\item When $m=\alpha=1.2$, the diameter is 0, which means positive pairs are pulled as close as possible and has the same effect as the conventional contrastive loss. In this case, the Recall@1 is considerably worse than the best. Especially, when $C=30, K =6$, without hypersphere regularisation, the Recall@1 is only 76.4\% while the best Recall@1 with strong regularisation is 79.1\%.      
	
	\item In both settings, we obtain almost the best performance when the hypersphere diameter is 0.8. In case where $C=30, K =6$, the Recall@$K$ results are more sensitive to the hypersphere diameter. 
	
\end{itemize}

%\subsection{Pending Training from scratch: without momentum, with hyper-parameters annealing along with training iterations}

%\subsection{Pending Different Architectures}

\subsection{Ablation Study on Other General Factors}

In this subsection, we present our empirical study on other method-independent factors of deep metric learning: 
%1) Single-level versus Multilevel embeddings; 
1) Batch size; 
2) Embedding size; 
3) Batch content.

%\subsubsection{Single-level versus Multilevel Embeddings}

%%%%%%%%%%%%%%%%%%%%%%%%%%%%%%%%
%\begin{table}[!h]
%	\caption{Single-level embeddings versus multilevel embedding on SOP in terms of Recall@$K$ (\%). L, M and H represent the low-level, mid-level  and high-level embedding respectively. (L,M,H) means the concatenation of low-level, mid-level and high-level embeddings. 
%		The corresponding embedding size is denoted by `-embedding size'. 
%	}
%	\label{table:feature_type}
%	\centering
%	\vspace{-2pt}
%	\begin{tabular}{lcccc}
%		\toprule
%		Embedding & R@1 & R@10 & R@100 \\
%		\midrule
%		L-512  &  76.1 & 88.8 & 94.9 \\
%		M-512  &  76.9 & 89.6 & 95.5 \\
%		H-512  & 76.1 & 89.1 & 95.4 \\
%		(L,M,H)-512$\times$3 & \textbf{79.8} & \textbf{91.3} & \textbf{96.3} \\
%		\midrule
%		L-1536  &  76.7 & 89.1 & 95.1 \\
%		M-1536  &  77.4 & 89.9 & 95.6 \\
%		H-1536  & 76.7 & 89.5 & 95.5 \\
%		%(L,M,H)-512$\times$3$\times$3 \\
%		\bottomrule
%	\end{tabular}
%	%\vspace{-0.3cm}
%\end{table}
%%%%%%%%%%%%%%%%%%%%%%%%%%%%%%%%%\\

%%%%%%%%%%%%%%%%%%%%%%%%%%%%%%%%
\begin{table}[!t]
	\caption{The results of different batch size on the SOP. 
	}
	\label{table:batch_size}
	\centering
	\setlength{\tabcolsep}{12pt}
	%\vspace{-2pt}
	\vspace{-0.1cm}
	\begin{tabular}{lcccc}
		\toprule
		Batch size & R@1 & R@10 & R@100 \\
		\midrule
		$120=40\times3$ & 79.2 & 90.9 & 96.2 \\
		
		$150=50\times3$ & 79.5 & 91.1 & 96.2 \\
		
		$165=55\times3$ & 79.7 & 91.2 & {96.3} \\
		
		$180=60\times3$ & {79.8} & {91.3} & {96.3} \\
		
		$195=65\times3$ & {79.8} & {91.3} & {96.3} \\
		
		\bottomrule
	\end{tabular}
\end{table}
%%%%%%%%%%%%%%%%%%%%%%%%%%%%%%%%%\\

%As demonstrated in Section~\ref{section:state_of_the_art}, we find that RLL performs better using the multilevel embedding. To compare different single-level embeddings with the multilevel embedding, we conduct experiments on SOP. The results are reported in Table~\ref{table:feature_type}. We observe that: 
%\vspace{-7pt}
%\begin{itemize}
%	\item All single-level embeddings perform worse than the multilevel embedding by around 3\%.
%	
%	%\vspace{-7pt}
%	\item The low-level embedding also performs very well in contrast with mid-level and high-level embeddings. The performance gap of different single-level embeddings is smaller than 1\%. We remark that the low-level embedding can be used for fast inference, which is indispensable for  resource-constrained computational devices, e.g., mobile phones.
%	
%	\item We note that single-level and multilevel embeddings have different embedding size. To make the comparison more fair and reasonable, we also present the performance of single-level embeddings with the same encoding size as the multilevel version. According to the observation in  Table~\ref{table:feature_type}, the superiority of multilevel embedding indeed comes from their complementary effect instead of the increase of embedding size.    
%\end{itemize}

\subsubsection{The Impact of Batch Size}

%%%%%%%%%%%%%%%%%%%%%%%%%%%%%%%%%\\

The batch size is usually important in deep metric learning. 
During training, we follow the few-shot retrieval setting, so that the batch size determines the scale of a problem we are going to solve every iteration. 
%As presented in Section~\ref{section:deep_model} and \ref{section:implemtation_details}, the batch size decides the number of negative classes in the gallery. 
We conduct experiments on the SOP to evaluate the influence of batch size in our approach. Specifically, we fix the number of images per class ($\forall c, Nc=K=3$) and only change the number of classes ($C \in \{40,50,55,60,65\}$) in each mini-batch. The results are reported in Table~\ref{table:batch_size}. 
We can see that RLL is not very sensitive to the batch size. 
% does not play a crucial role in . 
%The performance gap is only 0.6\% when $C$ changes from 120 to 195.

\subsubsection{The Impact of Embedding Size}
\label{ablation_study:embedding_size}

%%%%%%%%%%%%%%%%%%%%%%%%%%%%%%%%
\begin{table}[!h]
	\caption{The results of RLL-(L,M,H) with different embedding size on the SOP test dataset. 
		%(\textbf{To add:} 32x3, 16x3)
	}
	\label{table:embedding_size}
	\centering
	%\vspace{-2pt}
	\vspace{-0.1cm}
	\begin{tabular}{ccccc}
		\toprule
		Embedding size & R@1 & R@10 & R@100 \\
		\midrule
		$16\times 3$ & 61.8 & 80.8 & 91.9 \\
		$32\times 3$ & 73.8 & 88.0 & 94.9 \\
		$64\times 3$ & 77.9 & 90.3 & 95.8 \\
		$128\times 3$ & 79.2 & 91.0 & 96.2 \\
		$256\times 3$ & 79.5 & 91.1 & 96.2 \\
		$512\times 3$ & 79.8 & 91.3 & 96.3 \\
		$1024\times 3$ & 79.9 & 91.4 & 96.4 \\
		$1536\times 3$ & 80.2 & 91.4 & 96.4 \\
		\bottomrule
	\end{tabular}
\end{table}
%%%%%%%%%%%%%%%%%%%%%%%%%%%%%%%%%\\
%%%%%%%%%%%%%%%%%%%%%%%%%%%%%%%%
\begin{table}[!h]
	\caption{The results of different batch content $C\times K$ on the SOP. We fix $N=C\times K=180$ and change $C, K$.
	}
	\label{table:batch_content}
	\centering
	%\vspace{-2pt}
	\vspace{-0.1cm}
	\begin{tabular}{ccccc}
		\toprule
		Batch content ($N=C\times K$) & R@1 & R@10 & R@100 \\
		\midrule
		$N=180=90\times 2$ & 79.8 & 91.3 & 96.2 \\
		$N=180=60\times 3$ & {79.8} & {91.3} & {96.3} \\
		$N=180=60\times 4$ & {79.4} & {91.1} & {96.2} \\
		$N=180=36\times 5$ & {79.0} & {90.8} & {96.1} \\
		$N=180=30\times 6$ & {79.0} & {90.9} & {96.2} \\
		\bottomrule
	\end{tabular}
\end{table}
%%%%%%%%%%%%%%%%%%%%%%%%%%%%%%%%%\\
%%%%%%%%%%%%%%%%%%%%%%%%%%%%%%%%%\\

The feature dimension is another considerable factor when learning deep representations for downstream tasks. 
Generally, the objective is to encode an input into a low-dimensional feature vector so that the storage and computational complexity can be reduced on downstream tasks, e.g., fast image retrieval \cite{yu2018product,zhai2018making}.  
Therefore, in this subsection, we study the influence of embedding size in our RLL. In all experiments, we set $C=60, K=3, \alpha=1.2, m=0.4, T_n=10$.  
The results on the SOP are displayed in the Table~\ref{table:embedding_size}. We can see that generally a larger embedding size leads to a better performance. 
Finally, the performance increase becomes negligible. 
Therefore, due to the limited storage and faster computational speed requirement in practice, we can choose a smaller encoding size.
% e.g., $128\times 3$, whose performance is only 1\% lower than $1536\times 3$. 

\subsubsection{The Impact of Batch Content }

%%%%%%%%%%%%%%%%%%%%%%%%%%%%%%%%%\\

%Our proposed RLL learns deep discriminative representations iteratively on mini-batches, i.e., small-scale image retrieval tasks where every data point acts as the query while the remained data points serve as the gallery iteratively. 

%We compare the performance of multilevel representations, i.e., (L,M,H)-$512\times 3$.

In this subsection, we study the format of a few-shot retrieval task in every iteration, e.g., the number of classes $C$ and images per class $K$. To abstain the impact from batch size, we fix $N=C\times K=180$ and change $C, K$ at the same time.  
We set $\alpha=1.2, m=0.4, T_n=10$  in all experiments. 
%Concretely, we set $\alpha=1.2, m=0.4, T=10$ and evaluate the performance on SOP test dataset.  
We display the results of RLL-(L,M,H) in Table~\ref{table:batch_content}. 
We observe that when there are more classes and fewer images per class, i.e., a task becomes more difficult, we obtain better generalisation performance.

\subsection{Qualitative results}

\begin{figure}[!h]
	\centering
	\includegraphics[width=1.0\linewidth]{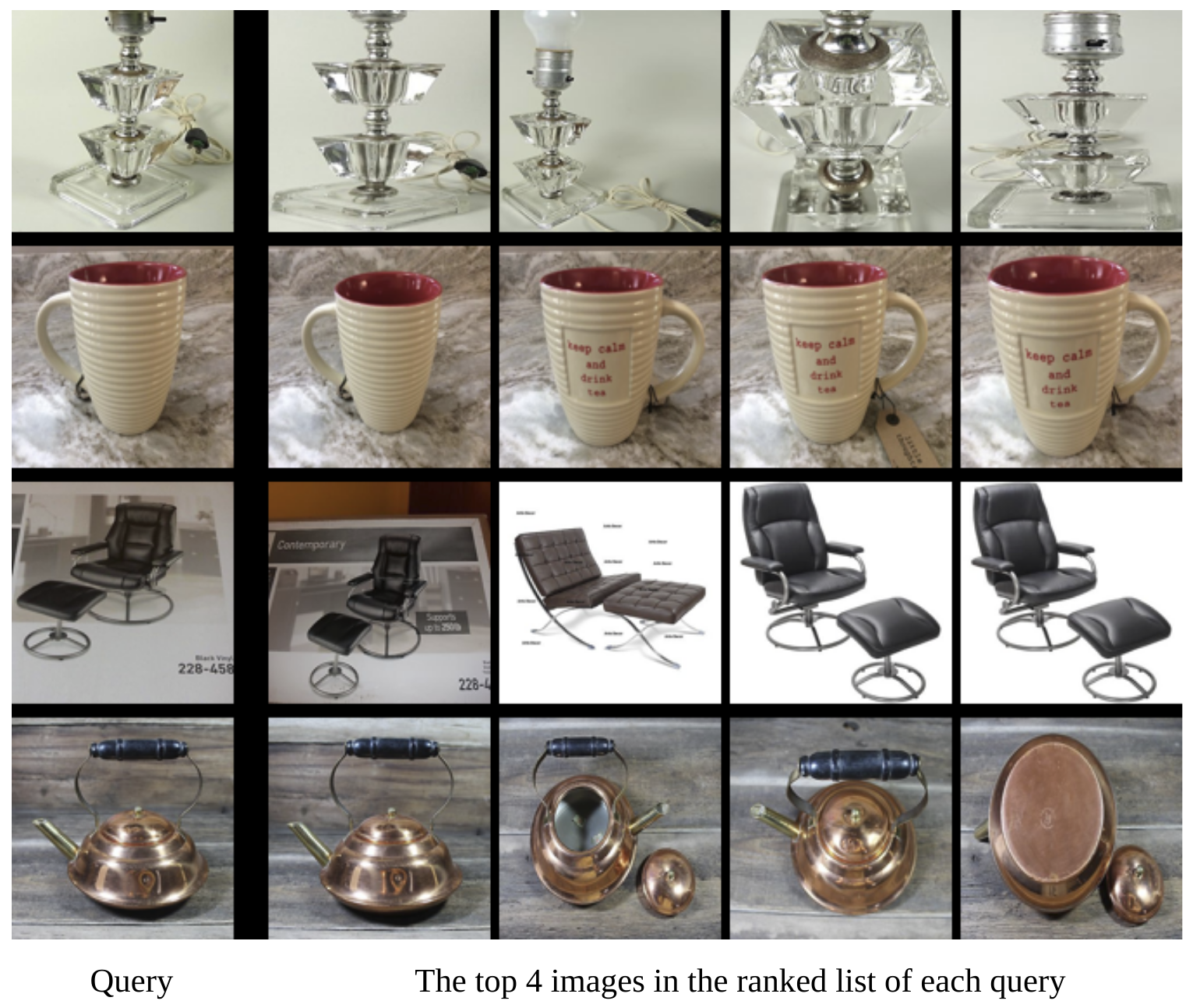}
	\caption{
		Visualisation of image retrieval on the SOP test dataset.  
		The leftmost column shows queries, which rank the images in the gallery according to the similarity.  
	}
	\label{fig:retrieval_visualisation}
\end{figure}

%\subsubsection{Visualisation of Image Retrieval}
\noindent\textbf{Visualisation of Image Retrieval.}
In Figure~\ref{fig:retrieval_visualisation}, we visualise the image retrieval results on the SOP test dataset. For every query, we show its top 4 images in the ranked list of the gallery set. We observe that the learned embedding model is robust and invariant to rotation and viewpoint. 

%\subsubsection{Visualisation of Image Clustering}

\begin{figure*}[!h]
    \centering
    \includegraphics[width=0.80\linewidth]{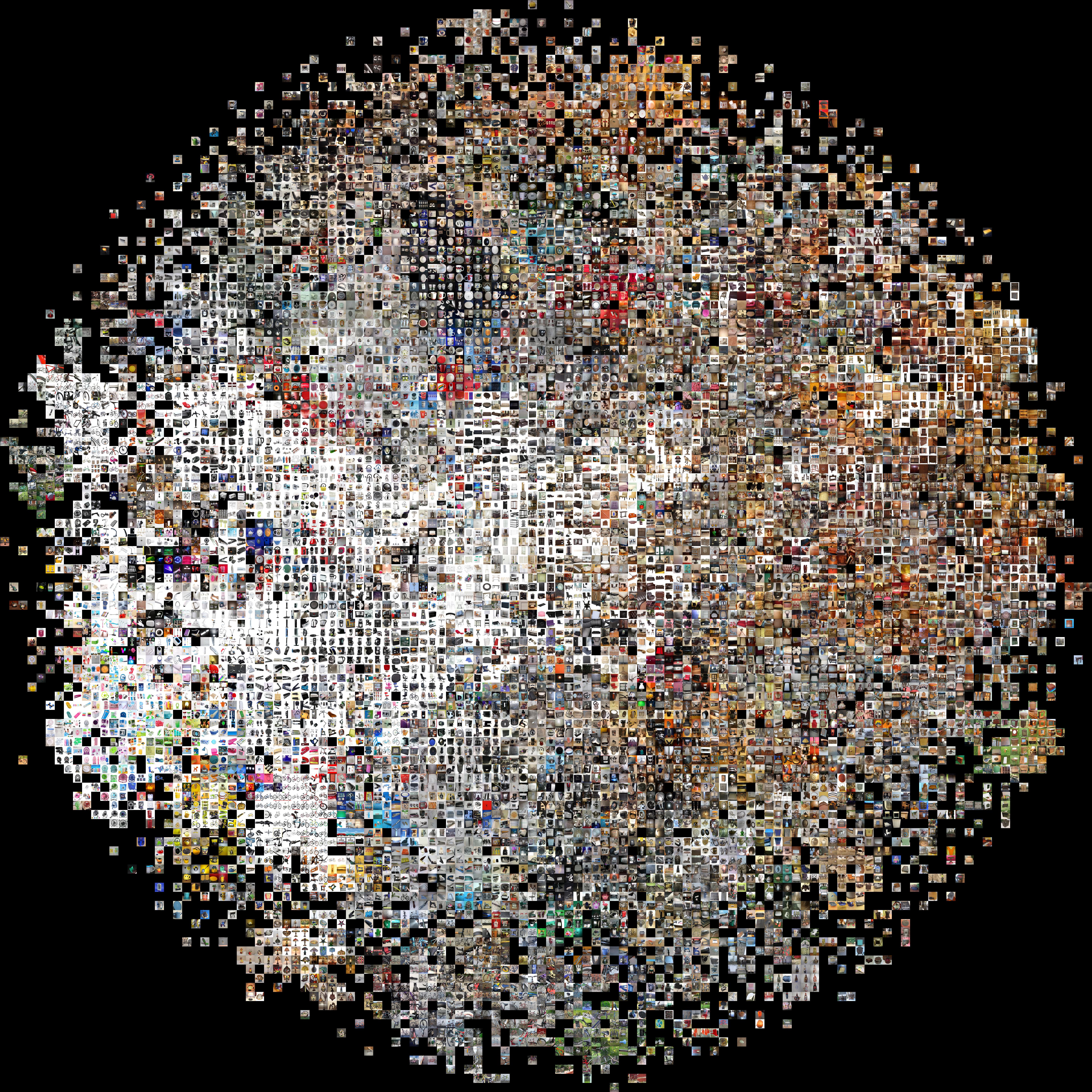}
    \caption{
        t-SNE visualisation~\cite{van2014accelerating} on the SOP test dataset. \textit{Best viewed on a monitor when zoomed in.}  
    }
    \label{fig:clustering_visualisation}
\end{figure*}

\noindent\textbf{Visualisation of Image Clustering.}
We visualise the image clustering result on the SOP test dataset in the Figure~\ref{fig:clustering_visualisation}.

%%%%%%%%%%%%%%%%%%%%%%%%%%%%%%%%
\begin{table}[!t]
	\caption{
		Study on the dynamic weighting scheme of negative pairs on the In-shop Clothes and CUB-200-2011 datasets. 
		We report the Recall@$1$ (\%) of every experiment. 
		The mean and standard deviation (SD) of two groups are shown in the bottom of each block.  
	}
	\label{table:dynamic_weighting_Tn}
	\centering
	\setlength{\tabcolsep}{10pt} % Default value: 6pt
	%\vspace{-6pt}
	%\fontsize{8.8pt}{9.8pt}\selectfont
	%
	\vspace{-0.1cm}
	\begin{tabular}{lcccc}
		\toprule
		\multirow{2}{*}{Period}
		&\multirow{2}{*}{\makecell{$T_1$}} & 
		\multirow{2}{*}{\makecell{$T_2$}} &
		
		%\multirow{2}{*}{\makecell{Hyperball \\Diameter\\$\alpha - m$}} & 
		%\multirow{2}{*}{\makecell{Emphasis\\Focus}} & \multirow{2}{*}{Model} & 
		%\multicolumn{1}{c}{\makecell{Recall@$1$}}
		\multirow{2}{*}{\makecell{In-shop Clothes}} & 
		\multirow{2}{*}{\makecell{CUB-200-2011}} 
		%& \multicolumn{2}{c}{ \makecell{Accuracy  on\\ Training Sets (\%)} }
		\\
		%\cmidrule{4-9}  
		& &  &  \\
		\midrule
        		\multirow{5}{*}{Early} & 4& \multirow{5}{*}{12} &86.2 & 53.2\\
        &8&& 87.3  & 55.0\\
        &12&& 87.7  & 56.1 \\
        &16&& 86.4  & 56.9\\
        &20&& 85.8  & 58.5\\
        
        \hdashline\\[-0.2cm]
        Mean$\pm \text{SD}$& & &86.68$\pm 0.627$&55.94$\pm 3.973$\\

        %\cmidrule{1-9}
		%\midrule
		%\cmidrule{2-5}
        \cmidrule{1-5}
       % \cmidrule{1-5}
		
	\multirow{5}{*}{Later}&
    \multirow{5}{*}{12}
    &4&87.7  & 56.0 \\
    &&8& 87.5  & 55.8 \\
    &&12& 87.7 & 56.1 \\
    &&16& 87.7 & 56.1 \\
    &&20& 87.7 & 56.4\\

    \hdashline\\[-0.2cm]
    Mean$\pm \text{SD}$ & & &87.66$\pm 0.008$&56.08$\pm 0.047$\\
		
		\bottomrule
	\end{tabular}
	%
	%\vspace{-0.3cm}
\end{table}
%%%%%%%%%%%%%%%%%%%%%%%%%%%%%%%%%

\section{Study on the critical learning periods of deep metric learning}
\label{section:critical_periods}

We have discussed in section~\ref{section:critical_learning_period} that the early learning period is critical in artificial deep neural networks. In this section, we study and validate them in the context of deep metric learning empirically from the perspective of example weighting.
%
%Our empirical study focuses on the dynamic negative examples weighting scheme ${T_n} = T_1 - cur\_iter * \frac{T_1-T_2}{max\_iter}$. $T_n$ changes along with training iterations. Our experimental settings are simple and symmetric as follows: 
%\begin{itemize}
%	\item To study whether the early learning phase is critical, we simply fix $T_2$ while comparing the results of different $T_1$.
%	\item Analogously, when studying the effect of the later learning phase, we simply fix $T_1$ while using different $T_2$. 
%\end{itemize}
%
Our results on the In-shop Clothes and CUB-200-2011 datasets are displayed in Table~\ref{table:dynamic_weighting_Tn}.
In these experiments, the batch size is 60. 
The number of training iterations is 30,000 on In-shop Clothes, while 10,000 on CUB-200-2011.
A significant observation is that \textit{$T_1$ is a more sensitive factor than $T_2$}: the performance's standard deviation of different $T_1$s is much larger than that of different $T_2$s.
Therefore, critical learning periods of deep metric learning also exist in the early learning phase. This is interesting and inspires us that more effort should be spent on the design of the early learning phase.  
%Moreover, it is generally argued that   

%\subsection{Study on The Early Learning Phase}
%\subsection{Exploration on The Later Learning Phase}

%\section{Application to Person Re-identification?}
%\label{section:person_reidentification_experiments}

%%%%%%%%%%%%%%%%%%%%%%%%%%%%%%%%%%%%%%%%
\section{Conclusion}
\label{section:conclusion}

In this paper, the ranked list loss is proposed to exploit all non-trivial data points in order to provide more informative supervision for learning discriminative embeddings. 
Following up our CVPR 2019 conference version, we further improve RLL to be a general extension of ranking-motivated losses. 
Concretely, given a query, RLL splits other data points into positive and negative sets, and forces a margin between them.
In addition, example mining and weighting are exploited to leverage all informative data points. 
Our proposed RLL achieves the state-of-the-art performance on two large datasets. 
Furthermore, we present many other interesting results, which are of high practical value and can be open leads of future research: 
(1) The results of RLL-L and RLL-M are highly competitive and even better although their network depth is shallower;  
(2) How to better exploit multi-layer embeddings in deep metric learning?   
(3) How to better design the early learning phase of deep metric leaning, since it is the critical learning stage.

% if have a single appendix:
%\appendix[Proof of the Zonklar Equations]
% or
%\appendix  % for no appendix heading
% do not use \section anymore after \appendix, only \section*
% is possibly needed

% use appendices with more than one appendix
% then use \section to start each appendix
% you must declare a \section before using any
% \subsection or using \label (\appendices by itself
% starts a section numbered zero.)
%

% you can choose not to have a title for an appendix
% if you want by leaving the argument blank
%\section{}
%Appendix two text goes here.

% use section* for acknowledgment
\ifCLASSOPTIONcompsoc
  % The Computer Society usually uses the plural form
  \section*{Acknowledgments}
\else
  % regular IEEE prefers the singular form
  %\section*{Acknowledgment}
\fi

%We thank AnyVision for offering Xinshao Wang University Special Research Scholarship during his PhD research at Queen's University Belfast, UK.

This work was supported in part by AnyVision Industrial Research Funding. 

%For Xinshao Wang, this work was mainly done when doing his PhD at Queen's University Belfast, which was funded by Anyvision Ltd.
%For  

%The authors would like to thank Anyvision Research Team, UK for the kind 

% Can use something like this to put references on a page
% by themselves when using endfloat and the captionsoff option.
\ifCLASSOPTIONcaptionsoff
  \newpage
\fi

% trigger a \newpage just before the given reference
% number - used to balance the columns on the last page
% adjust value as needed - may need to be readjusted if
% the document is modified later
%\IEEEtriggeratref{8}
% The "triggered" command can be changed if desired:
%\IEEEtriggercmd{\enlargethispage{-5in}}

% references section

% can use a bibliography generated by BibTeX as a .bbl file
% BibTeX documentation can be easily obtained at:
% http://mirror.ctan.org/biblio/bibtex/contrib/doc/
% The IEEEtran BibTeX style support page is at:
% http://www.michaelshell.org/tex/ieeetran/bibtex/
%\bibliographystyle{IEEEtran}
% argument is your BibTeX string definitions and bibliography database(s)
%\bibliography{IEEEabrv,../bib/paper}
%
% <OR> manually copy in the resultant .bbl file
% set second argument of \begin to the number of references
% (used to reserve space for the reference number labels box)
%\begin{thebibliography}{1}
%
%\bibitem{IEEEhowto:kopka}
%H.~Kopka and P.~W. Daly, \emph{A Guide to {\LaTeX}}, 3rd~ed.\hskip 1em plus
%  0.5em minus 0.4em\relax Harlow, England: Addison-Wesley, 1999.
%
%\end{thebibliography}

{%\small
	\bibliographystyle{ieee}
	\bibliography{ICML2020_DM,DML,DML_ICE}
}

% biography section
% 
% If you have an EPS/PDF photo (graphicx package needed) extra braces are
% needed around the contents of the optional argument to biography to prevent
% the LaTeX parser from getting confused when it sees the complicated
% \includegraphics command within an optional argument. (You could create
% your own custom macro containing the \includegraphics command to make things
% simpler here.)
%\begin{IEEEbiography}[{\includegraphics[width=1in,height=1.25in,clip,keepaspectratio]{mshell}}]{Michael Shell}
% or if you just want to reserve a space for a photo:

%\appendices
%\section{Visualisation of Image Clustering on SOP}
%
%We visualise the image clustering result on the SOP test dataset in Figure~\ref{fig:clustering_visualisation}.   

\vspace{-20pt}

%IEEEbiographynophoto 
%Use IEEEbiography if you would like to add a photo
\vspace{-0.1cm}
\begin{IEEEbiographynophoto}{Xinshao Wang} did his postdoc research at University of Oxford, UK after finishing his PhD at Queen's University Belfast, UK.
His PhD project was funded by University Special Research Scholarship, sponsored by AnyVision. 
Xinshao Wang has been working on core deep learning techniques with diverse applications: 
(1) Deep metric learning: to learn discriminative and robust image/video representations for downstream tasks, e.g., image/video retrieval and image/video clustering;
(2) Robust deep learning: robust learning and inference under adverse conditions, e.g., label noise, missing labels (semi-supervised learning), out-of-distribution training examples, sample imbalance, etc;
(3) Computer vision: video/set-based person re-identification, image/video classification/retrieval/clustering;
(4) AI health care.     
\end{IEEEbiographynophoto}

\vspace{-20pt}
% if you will not have a photo at all:
\begin{IEEEbiographynophoto}{Yang Hua}
is presently a lecturer at the Queen’s University of Belfast, UK. He received his Ph.D. degree from Universit\'e Grenoble Alpes / Inria Grenoble Rhne-Alpes, France, funded by Microsoft Research Inria Joint Center. He won PASCAL Visual Object Classes (VOC) Challenge Classification Competition in 2010, 2011 and 2012, respectively and the Thermal Imagery Visual Object Tracking (VOTTIR) Competition in 2015. His research interests include machine learning methods for image and video understanding. He holds three US patents and one China patent.
\end{IEEEbiographynophoto}

% insert where needed to balance the two columns on the last page with
% biographies
%\newpage

\vspace{-20pt}
\begin{IEEEbiographynophoto}{Elyor Kodirov}
%is currently a senior researcher at AnyVision, Belfast, UK. He
received his Ph.D. degree in the School of Electronic Engineering and Computer Science, Queen
Mary University of London, 2017 and the Master's degree in computer science from Chonnam National University, Korea, in 2014. His research interests include computer vision and machine learning.
\end{IEEEbiographynophoto}

\vspace{-20pt}
%IEEEbiographynophoto
%IEEEbiography
\begin{IEEEbiographynophoto}{Neil M. Robertson}
is Professor and Director of the Centre for Data Sciences and Scalable Computing, at the Queens University of Belfast, UK. He researches underpinning machine learning methods for visual analytics. His principal research focus is face and activity recognition in video. He started his career in the UK Scientific Civil Service with DERA (2000-2002) and QinetiQ (2002-2007). Neil was the 1851 Royal Commission Fellow at Oxford University (2003-2006) in the Robotics Research Group. His autonomous systems, defence and security research is extensive including UK major research programmes and doctoral training centres.
\end{IEEEbiographynophoto}

% You can push biographies down or up by placing
% a \vfill before or after them. The appropriate
% use of \vfill depends on what kind of text is
% on the last page and whether or not the columns
% are being equalized.

%\vfill

% Can be used to pull up biographies so that the bottom of the last one
% is flush with the other column.
%\enlargethispage{-5in}

% that's all folks
\end{document}